\documentclass{article}

\PassOptionsToPackage{numbers, compress}{natbib}
\usepackage[preprint]{neurips_2026}

\usepackage[utf8]{inputenc} 
\usepackage[T1]{fontenc}    
\usepackage{hyperref}       
\usepackage{url}            
\usepackage{booktabs}       
\usepackage{amsfonts}       
\usepackage{nicefrac}       
\usepackage{microtype}      
\usepackage[dvipsnames]{xcolor}

\usepackage{kotex}          
\usepackage{bbm}
\usepackage{amsmath}
\usepackage{amssymb}
\usepackage{graphicx}
\usepackage{mathtools}
\usepackage{multirow}
\usepackage{colortbl}
\usepackage{adjustbox}
\usepackage[normalem]{ulem}
\useunder{\uline}{\ul}{}

\usepackage{pdfrender}

\newcommand{\strokeBold}[1]{%
  \textpdfrender{TextRenderingMode=FillStroke,LineWidth=0.3pt}{#1}%
}

\usepackage{pifont}
\usepackage{xcolor}

\usepackage{placeins}
\usepackage{siunitx}

\usepackage{wrapfig}
\usepackage{lipsum}
\usepackage{siunitx}
\usepackage{textcomp}
\usepackage{subcaption}
\usepackage{enumitem}
\usepackage{float}
\usepackage{geometry}

\usepackage{algorithmic}
\usepackage{algorithm}
\usepackage{array}
\usepackage{bm}

\newcommand{\red}[1]{\textcolor{red}{#1}}
\newcommand{\blue}[1]{\textcolor{blue}{#1}}
\newcommand{\green}[1]{\textcolor{ForestGreen}{#1}}

\newcommand{\EGO}{\red{\textsc{Ego}}}
\newcommand{\AUXONE}{\green{\textsc{Aux1}}}
\newcommand{\AUXTWO}{\blue{\textsc{Aux2}}}

\usepackage{array}
\newcolumntype{L}[1]{>{\raggedright\let\newline\\\arraybackslash\hspace{0pt}}m{#1}}
\newcolumntype{C}[1]{>{\centering\let\newline\\\arraybackslash\hspace{0pt}}m{#1}}
\newcolumntype{R}[1]{>{\raggedleft\let\newline\\\arraybackslash\hspace{0pt}}m{#1}}

\title{Adaptation-Free Heterogeneous Collaborative Perception with Unseen Agent Configurations}

\author{%
  Hyunchul Bae\quad Heejin Ahn\thanks{Corresponding author.}\\
  School of Electrical Engineering\\
  Korea Advanced Institute of Science and Technology (KAIST)\\
  Daejeon, Korea 34141
  \texttt{\{bhc2675, heejin.ahn\}@kaist.ac.kr} \\
}

\begin{document}

\maketitle

\begin{abstract}
Collaborative perception improves 3D object detection by enabling agents to share complementary observations, but most existing methods assume fixed or known collaborator encoder configurations, limiting deployment in practice. 
In this work, we consider an open-world setting in which auxiliary agents with unseen configurations may appear after deployment, such as different LiDAR beam counts or encoder architectures. To address this challenge, we propose ALF, a collaborative perception framework that enables zero-adaptation collaboration with unseen agent configurations by lifting lightweight box-level messages into ego-compatible auxiliary features.
ALF converts auxiliary box-level messages into pseudo-BEV maps and synthesizes ego-compatible latent features by combining object-centric cues with scene context from the ego feature.
On V2X-Real, under a zero-shot evaluation across 64 case studies, ALF outperforms the strongest prior baseline by 35.91\% in relative mAP@0.7 
while requiring only 120 bytes per agent per frame ($\approx$ 9.6 Kbps bandwidth at 10 Hz). 
\end{abstract}

\section{Introduction}

\label{sec:intro}
Collaborative perception enables multiple agents to exchange information and jointly reason about a scene beyond the sensing limits of any individual agent. By aggregating complementary observations, it improves spatial coverage and reduces uncertainty caused by occlusion and limited sensing range~\cite{huang2025cp-review1, han2023cp-review2, gao2024cp-review3}. Autonomous driving is a prominent application of this paradigm, where vehicles and infrastructure communicate to jointly perceive the environment~\cite{wang2025vicooper, bae2024rethinkingroleinfra, gao2025airv2x}. 

Collaborative perception frameworks are commonly categorized as early, intermediate, and late fusion, depending on the type of information exchanged: raw sensor data, latent feature representations, and object-level predictions.
Among these, intermediate fusion strikes a balance between communication cost and representational fidelity~\cite{xu2022cobevt, xu2022v2xvit, hu2022where2comm, lu2023coalign, zhang2024ermvp}. It preserves richer scene context than late fusion while avoiding the bandwidth overhead of early fusion.
However, its central challenge is representation heterogeneity: features from auxiliary agents may be incompatible with the ego representation due to differences in \emph{agent configurations}. We define an agent configuration as the combination of sensor specifications and encoder configurations, covering hardware-dependent properties such as LiDAR beam count and model-dependent choices such as backbone architecture, voxelization settings, and feature resolution.

\begin{figure} [t!]
    \centering
    \includegraphics[width=1.0\linewidth]{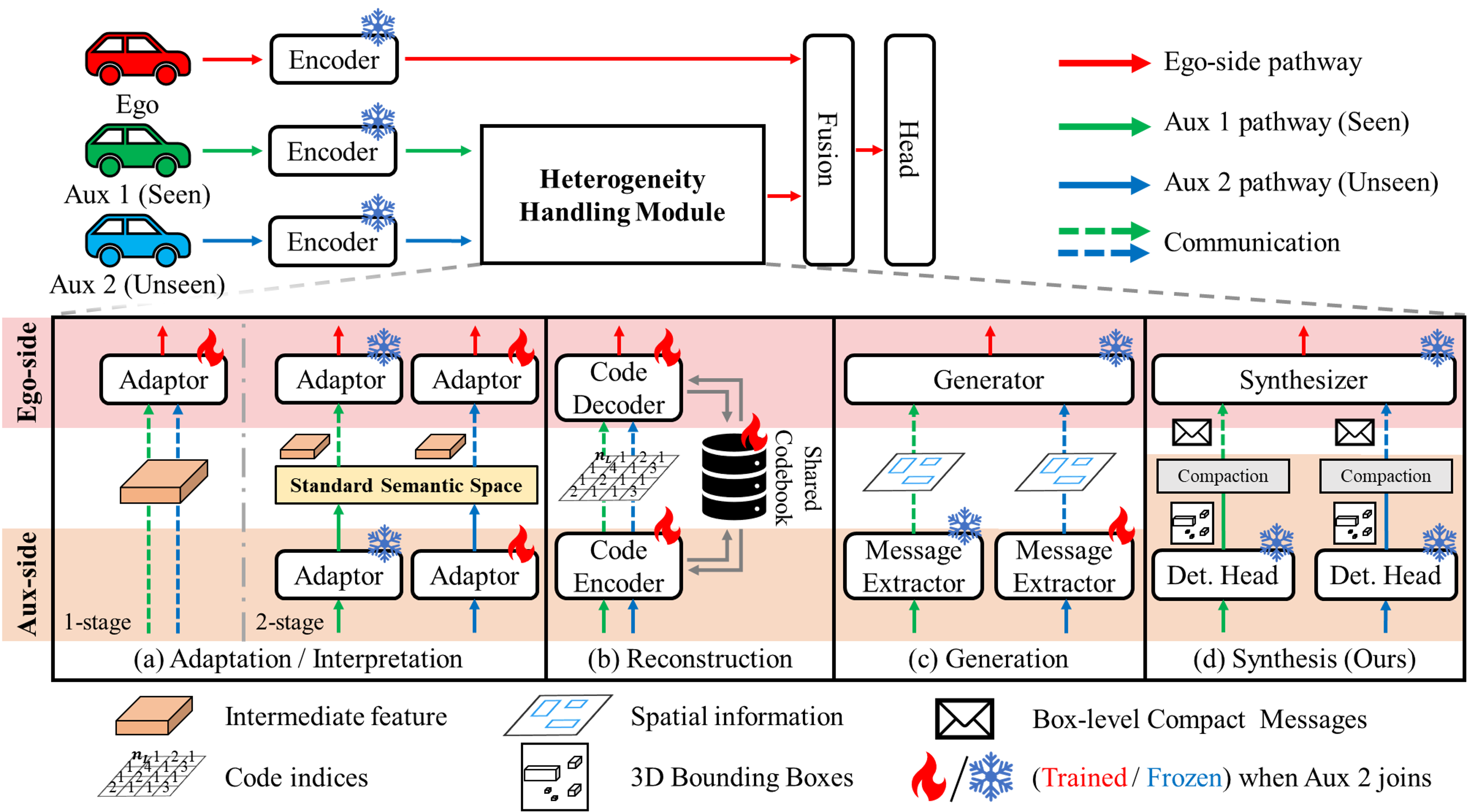}
    \caption{Comparison of strategies for integrating an unseen auxiliary agent (\AUXTWO) in heterogeneous collaborative perception, after fine-tuning with a seen auxiliary agent (\AUXONE):
(a) requires training additional adaptors for \AUXTWO,
(b) requires updating the shared codebook and code encoder/decoder,
(c) requires training the message extractor of \AUXTWO,
while (d) requires no additional training.}\label{fig:sec1_introduction}
    \vspace{-15pt}
\end{figure}

In realistic deployments, heterogeneity must be handled in an open-world setting, where an ego agent must collaborate with previously unseen auxiliary agents whose agent configurations were not observed during training. We refer to this scenario as \emph{configuration-agnostic heterogeneity}. This setting is challenging because independently developed agents do not, in general, share a common latent interface. In contrast, most prior work addresses heterogeneity only within a closed set of agent configurations observed during training or fine-tuning, which we refer to as \emph{configuration-specific heterogeneity}.


Prior heterogeneous intermediate-fusion methods mainly rely on adaptation \cite{lu2024heal, xu2023mpda, xia2025polyinter, luo2024pnpda, gao2025stamp}, reconstruction \cite{hu2024codefilling}, or generation \cite{zhou2025gencomm}, as illustrated in Fig.~\ref{fig:sec1_introduction}. These approaches typically require additional training to accommodate previously unseen auxiliary encoder configurations. This need for repeated per-agent retraining limits scalability and can degrade fusion performance when received representations fall outside the training distribution.

We propose \textbf{Adaptation-Free Late-to-Intermediate Fusion (ALF)}, a new route to collaborative perception that avoids transmitting configuration-dependent latent features. Instead, each auxiliary agent sends a compact box-level message derived from its predicted 3D boxes, and the ego agent synthesizes an ego-compatible feature from this object-centric cue before fusion (Fig.~\ref{fig:sec1_introduction}(d)). 
The key idea of ALF is to decouple communication and fusion: auxiliary information is communicated through a late-fusion-style interface, yet fused in an intermediate-fusion manner. Box-level messages provide a stable communication interface across heterogeneous auxiliary agents. However, direct late fusion is limited, as such detections can only be used at the final decision stage. ALF therefore lifts these object-centric cues into a compatible latent representation.
This lifting is guided by the ego latent feature, which provides scene context to transform object-centric cues into a fusion-ready representation. In this way, ALF recovers the pre-detection interaction benefits of intermediate fusion without requiring alignment of arbitrary auxiliary latent spaces. As a result, ALF enables plug-and-play collaboration with heterogeneous auxiliary agents—including those with previously unseen agent configurations—under a sub-1KB communication budget, as validated on V2X-Real~\cite{xiang2024v2xreal}.



We summarize our contributions as follows.
\begin{itemize}[topsep=0pt, noitemsep]
    \item[\textbullet] We propose ALF, an adaptation-free late-to-intermediate fusion framework that lifts box-level messages into ego-compatible latent features using ego latent features, enabling plug-and-play collaboration with heterogeneous auxiliary agents.
    \item[\textbullet] We solve configuration-agnostic heterogeneity, an open-world setting in which unseen auxiliary agent configurations may appear at inference time.
    \item[\textbullet] We demonstrate that ALF outperforms the strongest prior baseline on V2X-Real under unseen agent configurations, while requiring only a sub-1KB communication budget without any configuration-specific adaptation.
\end{itemize}


\section{Related Work}
\label{sec:related_works}
\paragraph{Collaborative Perception Methods}
Intermediate fusion has been the main focus in collaborative perception due to its favorable accuracy-bandwidth trade-off. F-Cooper~\cite{chen2019fcooper} uses maxout-based feature aggregation. V2X-ViT~\cite{xu2022v2xvit} introduces heterogeneous multi-agent attention and multi-scale window attention. CoBEVT~\cite{xu2022cobevt} adopts fused axial attention for efficient local-global aggregation. Where2comm~\cite{hu2022where2comm} reduces communication cost by transmitting sparse foreground regions. MRCNet~\cite{hong2024mrcnet} further improves robustness in noisy environments through motion-aware communication. A related line uses bounding boxes together with intermediate fusion in two ways. Some methods use boxes as geometric anchors for feature alignment, as in CoAlign~\cite{lu2023coalign} and RoCo~\cite{huang2024roco}, which correct relative poses from matched object detections before feature fusion. Others use boxes as auxiliary signals, as in CoSDH~\cite{xu2025cosdh}, which applies confidence-aware late fusion on detection boxes to refine intermediate-fusion results under low bandwidth. In contrast, ALF uses compact box-level messages as the primary collaboration interface and lifts them into ego-compatible latent representations, enabling feature-level fusion without relying on configuration-dependent intermediate features.

\paragraph{Heterogeneity Handling Methods}
Heterogeneous encoders introduce a domain gap between ego and auxiliary features, making direct intermediate fusion challenging. Existing approaches can be broadly categorized into three types.
The first is the adaptation- or interpretation-based approach, as shown in Fig.~\ref{fig:sec1_introduction}(a). This approach learns mappings either to the ego feature domain or to a shared semantic space. For example, MPDA~\cite{xu2023mpda} learns a feature resizer and a sparse cross-domain transformer. HEAL~\cite{lu2024heal} builds a unified collaboration space and performs backward alignment by freezing the ego-side fusion and detection head. PnPDA~\cite{luo2024pnpda} and STAMP~\cite{gao2025stamp} emphasize non-intrusive and scalable alignment. PolyInter~\cite{xia2025polyinter} uses a prompt-driven interpreter.
The second is the reconstruction-based approach, as shown in Fig.~\ref{fig:sec1_introduction}(b). Rather than transmitting latent features, methods such as CodeFilling~\cite{hu2024codefilling} send compact code indices and reconstruct latent features from a shared codebook.
The third is the generation-based approach, as shown in Fig.~\ref{fig:sec1_introduction}(c). Methods such as GenComm~\cite{zhou2025gencomm} generate collaborator-aligned features from spatial cues, requiring only lightweight extractor tuning for newly introduced agents. 
In contrast, ALF is distinct from all three categories. ALF adopts a synthesis-based design that lifts auxiliary box-level messages into the ego-compatible feature by leveraging ego latent features. The resulting representations are directly compatible for fusion, eliminating the need for retraining when new auxiliary agents are introduced.

\section{Methodology}
\label{sec:method}
\subsection{Problem Formulation}
\label{subsec:problem_formulation}
We consider collaborative 3D object detection with one ego agent and a time-varying set of auxiliary agents. Throughout this paper, we reserve the index $1$ for the ego agent. The ego agent observes sensor data $\mathcal{P}_1$ and extracts an ego latent feature $\mathcal{F}_1$. Each auxiliary agent $i \in \mathcal{A}_{\mathrm{aux}}$ has its own agent configuration $c_i \in \mathcal{C}$ and sends a message $\mathcal{M}_{i\rightarrow 1}$ to the ego agent. Here, an agent configuration refers to the attributes that determine the latent feature domain, including sensor specifications and encoder configurations.

Let $\mathcal{C}_{\mathrm{seen}} \subset \mathcal{C}$ denote the set of auxiliary agent configurations observed during training. We study an open-world deployment setting in which the auxiliary agent configuration set is not fixed after training. The deployment-time auxiliary set $\mathcal{A}_{\mathrm{aux}}$ includes unseen agent configurations, i.e., there exists an agent $i\in\mathcal{A}_{\mathrm{aux}}$ such that $c_i \notin \mathcal{C}_{\mathrm{seen}}$. 

Our goal is to learn a heterogeneity-handling module $\mathcal{H}_{\theta}$ that resolves configuration-agnostic heterogeneity while maintaining strong detection performance under a communication budget. Given the ego latent feature $\mathcal{F}_1$ and the auxiliary messages $\{\mathcal{M}_{i\rightarrow 1}\}_{i\in\mathcal{A}_{\mathrm{aux}}}$, we formulate the problem as
\begin{equation*}
\begin{aligned}
\theta^\star
=
\arg\max_{\theta}\;&
\mathcal{J}\!\left(
\mathcal{B}_1^\star,\;
\mathcal{CP}\!\left(
\mathcal{F}_1,\,
\mathcal{H}_{\theta}\!\left(
\{\mathcal{M}_{i\rightarrow 1}\}_{i\in\mathcal{A}_{\mathrm{aux}}}
\right)
\right)
\right) \\
\text{subject to}\;&
\sum_{i\in\mathcal{A}_{\mathrm{aux}}}
b(\mathcal{M}_{i\rightarrow 1})
\le B,
\quad
\exists\, i\in\mathcal{A}_{\mathrm{aux}}
\ \text{such that}\ 
c_i \notin \mathcal{C}_{\mathrm{seen}}.
\end{aligned}
\end{equation*}
where $\mathcal{B}_1^\star$ denotes the ground-truth 3D bounding boxes, 
$\mathcal{J}$ denotes a 3D object detection evaluation function, such as average precision (AP), $\mathcal{CP}$ denotes the collaborative perception detector, which consists of a feature fusion module and a detection head and produces the final prediction, $b(\cdot)$ measures communication cost, and $B$ is the communication budget.


\begin{figure}[t!]
    \centering
    \includegraphics[width=0.99\linewidth]{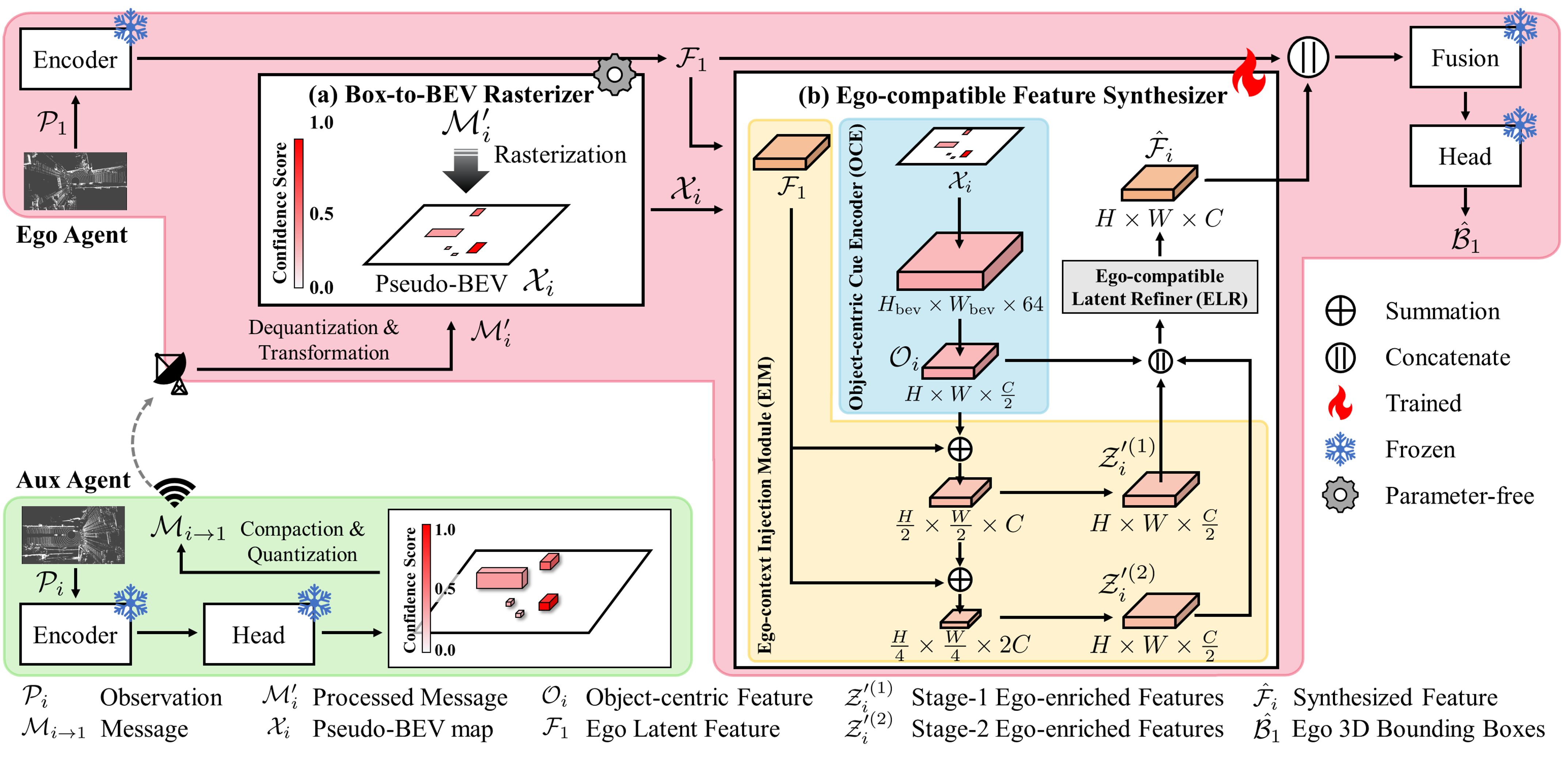}
    \caption{Overall architecture of ALF. (a) The box-level message is converted into a pseudo-BEV map by the Box-to-BEV Rasterizer (B2BR).
(b) The Ego-compatible Feature Synthesizer (EFS) lifts the sparse pseudo-BEV cue into an ego-compatible latent feature using the ego latent feature.}
    \label{fig:sec3_overall_architecture}
\end{figure}

\subsection{Method Overview}
\label{subsec:method_overview}

ALF enables late-fusion-style communication and intermediate-style feature fusion for open-world heterogeneous collaborative perception. Instead of exchanging configuration-dependent intermediate features, each auxiliary agent sends a compact box-level message, which the ego agent lifts into the ego latent space. This design preserves the interoperability and low communication cost of late fusion while enabling pre-detection interaction within a frozen ego-stack.

The overall pipeline has three stages. First, each auxiliary agent converts its local detections into a compact box-level message $\mathcal{M}_{i\rightarrow 1}$ through field selection and quantization.
Second, the ego agent dequantizes and transforms the received message into the ego frame using the relative pose $T_{i\rightarrow 1}$ and converts it into a high-resolution pseudo-BEV cue $\mathcal{X}_i$ with the Box-to-BEV Rasterizer (B2BR). 
Third, the Ego-compatible Feature Synthesizer (EFS) integrates $\mathcal{X}_i$ with the ego latent feature $\mathcal{F}_1\in\mathbb{R}^{H\times W\times C}$ to lift a fusion-ready collaborative feature $\hat{\mathcal{F}}_i\in\mathbb{R}^{H\times W\times C}$.

The heterogeneity-handling module $\mathcal{H}_{\theta}$ introduced in Section~\ref{subsec:problem_formulation} is instantiated as a sequential composition of dequantization, coordinate transformation, B2BR, and EFS:
\begin{equation*}
\hat{\mathcal{F}}_i
=
\mathcal{H}_{\theta}\!\left(\mathcal{M}_{i\rightarrow 1};\,T_{i\rightarrow 1},\,\mathcal{F}_1\right)
=
f_{\mathrm{EFS},\theta}\!\left(
f_{\mathrm{B2BR}}\!\left(
\mathcal{T}\!\left(
\mathcal{D}\!\left(\mathcal{M}_{i\rightarrow 1}\right),
T_{i\rightarrow 1}
\right)
\right),
\mathcal{F}_1
\right),
\end{equation*}
where $f_{\mathrm{EFS},\theta}$ denotes the EFS, $f_{\mathrm{B2BR}}$ denotes the B2BR, $\mathcal{T}$ denotes transformation into the ego frame using $T_{i\rightarrow1}$, and $\mathcal{D}$ denotes dequantization. After synthesizing the ego-compatible feature $\hat{\mathcal{F}}_i$, the final 3D bounding boxes $\hat{\mathcal{B}}_1$are predicted through the collaborative perception model $\mathcal{CP}$:
\begin{equation*}
\hat{\mathcal{B}}_1=\mathcal{CP}\left(\mathcal{F}_1,\{\hat{\mathcal{F}}_i\}_{i\in\mathcal{A}_{\mathrm{aux}}}\right).
\end{equation*}
Fig.~\ref{fig:sec3_overall_architecture} summarizes the full pipeline.

\subsection{Box-Level Message Compaction}
\label{subsec:message_and_projection}

ALF communicates through box-level messages rather than configuration-dependent latent features. This choice is central to our open-world setting, where auxiliary agents may use unseen agent configurations and therefore cannot be assumed to share a compatible latent space with the ego agent.

Each auxiliary agent $i\in \mathcal{A}_{\mathrm{aux}}$ sends a message defined as $
\mathcal{M}_{i\rightarrow 1}=\{m_{i,k}\}_{k=1}^{K_{\max}}$ 
where $K_{\max}$ denotes the maximum number of prediction boxes and $m_{i,k}$ contains the BEV-relevant fields of a prediction box.
In particular, we define $m_{i,k}=(\tilde{x}_{i,k},\tilde{y}_{i,k},\tilde{w}_{i,k},\tilde{\ell}_{i,k},\tilde{\psi}_{i,k},\tilde{s}_{i,k})$
where $\tilde{x}_{i,k}$ and $\tilde{y}_{i,k}$ denote the box center in the local BEV plane, $\tilde{w}_{i,k}$ and $\tilde{\ell}_{i,k}$ denote the box width and length, $\tilde{\psi}_{i,k}$ denotes the yaw angle, and $\tilde{s}_{i,k}$ denotes the confidence score. Here, $\tilde{(\cdot)}$ indicates quantized values. Quantization details are provided in Appendix~\ref{app:subsec:quantization_details}.

If the detector produces fewer than $K_{\max}$ boxes, the remaining entries are zero-padded. If it produces more, we retain the top-$K_{\max}$ boxes ranked by confidence. 
This compact representation preserves essential object evidence for collaboration while reducing communication cost compared to conventional late fusion.

\subsection{Box-to-BEV Rasterizer}
\label{subsec:b2br}
We now convert the compact box-level message into a pseudo-BEV representation. After transmission, the ego agent first dequantizes the received message and transforms the spatial fields (position, size, orientation) into the ego frame using relative pose $T_{i\rightarrow 1}$, yielding a processed message $\mathcal{M}'_i$.
We then convert $\mathcal{M}'_i$ into a pseudo-BEV map $\mathcal{X}_i$ with the B2BR:
\[
\mathcal{X}_i=f_{\mathrm{B2BR}}(\mathcal{M}'_i)\in\mathbb{R}^{H_{\mathrm{bev}}\times W_{\mathrm{bev}}\times 1},
\]
where $H_{\mathrm{bev}}$ and $W_{\mathrm{bev}}$ denote the height and width of the BEV grid.

This pseudo-BEV representation $\mathcal{X}_i$ serves two purposes. First, it preserves the spatial footprint of auxiliary detections in a shared representation that is independent of the auxiliary agent configuration. Second, it provides a high-resolution object-centric cue that can be lifted into the ego latent space by the subsequent synthesizer. Additional details of B2BR, including the exact definitions of $H_{\mathrm{bev}}$ and $W_{\mathrm{bev}}$, are provided in Appendix~\ref{app:subsec:b2br_details}.

\subsection{Ego-compatible Feature Synthesizer}
\label{subsec:efs}
The pseudo-BEV map $\mathcal{X}_i$ is a sparse representation of object-level information. 
To achieve intermediate-fusion-style interaction, we introduce the EFS to lift the pseudo-BEV representation into the ego latent space:
$$\hat{\mathcal{F}}_i=f_{\mathrm{EFS},\theta}(\mathcal{X}_i,\mathcal{F}_1)\in\mathbb{R}^{H\times W\times C}.$$
Here, ``ego-compatible'' means compatibility with the latent space expected by the frozen ego fusion and detection modules, rather than reconstruction of the auxiliary agent's native feature space.

EFS consists of three blocks:
\[
\mathcal{O}_i=\Phi_{\mathrm{OCE}}(\mathcal{X}_i),\qquad
(\mathcal{Z}_i'^{(1)},\mathcal{Z}_i'^{(2)})=\Phi_{\mathrm{EIM}}(\mathcal{O}_i,\mathcal{F}_1),\qquad
\hat{\mathcal{F}}_i=\Phi_{\mathrm{ELR}}\!\left([\mathcal{O}_i,\mathcal{Z}_i'^{(1)},\mathcal{Z}_i'^{(2)}]\right).
\]

The \textbf{Object-centric Cue Encoder} (OCE) (blue region in Fig.~\ref{fig:sec3_overall_architecture}(b)) converts the pseudo-BEV map $\mathcal{X}_i$ into a compact object-centric feature $\mathcal{O}_i$. It reduces spatial resolution while increasing the channel dimension to match the scale of the ego latent feature. At the same time, it extracts object-level information from the pseudo-BEV map, making $\mathcal{O}_i$ suitable for subsequent synthesis:
\begin{equation*}
\mathcal{O}_i = \mathrm{Enc}_{L}(\mathrm{Expand}(\mathcal{X}_i)) \in \mathbb{R}^{H\times W\times C/2},
\end{equation*}
where $\mathrm{Expand}(\cdot)$ denotes shallow channel expansion of the pseudo-BEV map $\mathcal{X}_i$ to the channel dimension of 64 in our setting. $\mathrm{Enc}_{L}(\cdot)$ denotes an $L$-stage encoder built from Conv--BN--SiLU blocks, which progressively reduces the spatial resolution while gradually increasing the channel dimension.

The \textbf{Ego-context Injection Module} (EIM) (yellow region in Fig.~\ref{fig:sec3_overall_architecture}(b)) enriches the object-centric feature $\mathcal{O}_i$ with the ego latent feature $\mathcal{F}_1$, which provides scene geometry and semantics. EIM follows a two-stage pyramid-style design that enables multi-scale context integration. In the first stage, the ego latent feature is added to $\mathcal{O}_i$ before downsampling. In the second stage, the ego latent feature is added again to the first-stage feature before a further downsampling step. The resulting two-stage features are then upsampled to the resolution of $\mathcal{F}_1$, yielding
\begin{align*}
\mathcal{Z}_i'^{(1)}
&=
\mathrm{Up}_{1}\!\left(
\mathrm{Down}_{1}\!\left(
\mathcal{O}_i+\mathrm{AvgPool}(\mathcal{F}_1)
\right)
\right)
\in
\mathbb{R}^{H\times W\times C/2}, \\
\mathcal{Z}_i'^{(2)}
&=
\mathrm{Up}_{2}\!\left(
\mathrm{Down}_{2}\!\left(
\mathrm{Down}_{1}\!\left(\mathcal{O}_i\right)+\mathrm{AvgPool}(\mathcal{F}_1)
\right)
\right)
\in
\mathbb{R}^{H\times W\times C/2}.
\end{align*}
Here, $\mathrm{AvgPool}(\cdot)$ denotes average pooling of $\mathcal{F}_1$ to match the spatial resolution of the feature it is added to. $\mathrm{Down}_{1}(\cdot)$ and $\mathrm{Down}_{2}(\cdot)$ denote the first- and second-stage downsampling operators, respectively, while $\mathrm{Up}_{1}(\cdot)$ and $\mathrm{Up}_{2}(\cdot)$ denote the corresponding upsampling operators that first restore spatial resolution via bilinear upsampling and then apply Conv--BN--SiLU blocks.


The \textbf{Ego-compatible Latent Refiner} (ELR) (gray region in Fig.~\ref{fig:sec3_overall_architecture}(b)) synthesizes the final fusion-ready feature by aggregating the object-centric feature $\mathcal{O}_i$ with the ego-enriched features $\mathcal{Z}_i'^{(1)}$ and $\mathcal{Z}_i'^{(2)}$. Although these inputs provide complementary local information and broader scene context, they are not yet aligned with the latent space expected by the ego fusion module.

ELR first projects the concatenated feature to channel dimension $C$ using two convolutional blocks, and then refines the projected feature with three residual blocks while preserving the backbone resolution. The final output is the ego-compatible latent feature $\hat{\mathcal{F}}_i$:
\begin{align*}
\tilde{\mathcal{F}}_i
=
\mathrm{Proj}\!\left(
[\mathcal{O}_i,\mathcal{Z}_i'^{(1)},\mathcal{Z}_i'^{(2)}]
\right)
\in
\mathbb{R}^{H\times W\times C}, &&
\hat{\mathcal{F}}_i
=
\mathrm{ResBlock}_{3}\!\left(
\tilde{\mathcal{F}}_i
\right)
\in
\mathbb{R}^{H\times W\times C}.
\end{align*}
Here, $\mathrm{Proj}(\cdot)$ denotes projection to channel dimension $C$, and $\mathrm{ResBlock}_{3}(\cdot)$ denotes a stack of three residual blocks that refines the latent representation.
Detailed architectural configurations are provided in Appendix~\ref{app:subsec:EFS_Architecture}.

\subsection{Loss}
\label{subsec:training_objective}

We train ALF using two objectives: a standard ego detection loss $\mathcal{L}_{\mathrm{det}}$ and an auxiliary feature-alignment loss $\mathcal{L}_{\mathrm{align}}$. The overall objective is
\[
\mathcal{L}
=
\alpha_{\mathrm{det}}\mathcal{L}_{\mathrm{det}}
+
\alpha_{\mathrm{align}}
\frac{1}{|\mathcal{A}_{\mathrm{aux}}|}
\sum_{i\in\mathcal{A}_{\mathrm{aux}}}
\mathcal{L}_{\mathrm{align}}^{(i)}.
\]
The detection loss supervises the final output using the same objective as the frozen ego detector: $\mathcal{L}_{\mathrm{det}}
=
\alpha_{\mathrm{cls}}\,\mathcal{L}_{\mathrm{cls}}
+
\alpha_{\mathrm{reg}}\,\mathcal{L}_{\mathrm{reg}},$
where $\mathcal{L}_{\mathrm{cls}}$ is the focal classification loss~\cite{lin2017focalloss} and $\mathcal{L}_{\mathrm{reg}}$ is the Smooth L1 box regression loss~\cite{girshick2015smoothl1}.

For each auxiliary agent $i$, we align the synthesized feature $\hat{\mathcal{F}}_i$ to an ego-domain teacher feature $\mathcal{F}_i^\star$. This teacher feature is obtained by transforming the auxiliary observation into the ego frame and encoding it with the frozen ego encoder. Since foreground objects occupy only a small fraction of the BEV space, a global alignment would be dominated by background areas. To address this, we compute the alignment loss separately for object and background regions:
\[
\mathcal{L}_{\mathrm{align}}^{(i)}
=
\alpha_{\mathrm{obj}}\,
\mathcal{L}_{\mathrm{cos}}(\hat{\mathcal{F}}_i,\mathcal{F}_i^\star;\mathbf{\Omega}_i^{\mathrm{obj}})
+
\alpha_{\mathrm{bg}}\,
\mathcal{L}_{\mathrm{cos}}(\hat{\mathcal{F}}_i,\mathcal{F}_i^\star;\mathbf{\Omega}_i^{\mathrm{bg}}),
\]
where $\mathbf{\Omega}_i^{\mathrm{obj}}$ and $\mathbf{\Omega}_i^{\mathrm{bg}}$ denote the object and background masks, respectively.
Given a region mask $\mathbf{\Omega}$, the region-weighted cosine alignment loss is defined as
\[
\mathcal{L}_{\mathrm{cos}}(\hat{\mathcal{F}}_i,\mathcal{F}_i^\star;\mathbf{\Omega})
=
\frac{1}{\max(1,\|\mathbf{\Omega}\|_1)}
\sum_{u=1}^{H}\sum_{v=1}^{W}
\mathbf{\Omega}(u,v)
\left(1-\cos\!\left(\hat{\mathcal{F}}_i(u,v),\mathcal{F}_i^\star(u,v)\right)\right).
\]

This objective encourages the synthesized feature to match the ego-space teacher while ensuring that object-level information is preserved despite the dominance of background regions. Detailed constructions of the teacher features and masks are provided in Appendix~\ref{app:subsec:teacher_feature} and \ref{app:subsec:pseudo_bev_mask}, respectively.

\subsection{Training Strategy and Adaptation-Free Deployment}
\label{subsec:training_and_deployment}

ALF is trained in a single fine-tuning stage. During training, only the parameters $\theta$ of EFS are optimized, while the pretrained ego encoder, fusion module, and detection head remain frozen. Auxiliary agents are treated as fixed black-box detectors.
Once trained, EFS remains fixed during deployment. When auxiliary agents with unseen agent configurations are encountered at inference time, ALF requires no additional fine-tuning or retraining. Instead, any auxiliary agent can immediately participate in collaboration by providing a box-level message $\mathcal{M}_{i\rightarrow 1}$ and relative pose $T_{i\rightarrow 1}$.

Because ALF interfaces with auxiliary agents exclusively through this universal box-level message format, it enables plug-and-play, adaptation-free deployment in open-world settings.


\section{Experiments}
\label{sec:experiments}
\subsection{Experiments Setup}
\label{subsec:exper_setting}
\paragraph{Dataset, metrics, and baselines.}
We evaluate ALF on V2X-Real~\cite{xiang2024v2xreal}, a large-scale real-world V2X cooperative perception dataset. Performance is measured by Average Precision (AP) for each superclass, and we report the mean Average Precision (mAP) across all superclasses. Specifically, we report mAP@0.5 and mAP@0.7, corresponding to IoU thresholds of 0.5 and 0.7, respectively. 

To evaluate ALF under heterogeneous settings, we compare it with representative baselines, including E2E, MPDA~\cite{xu2023mpda}, CodeFilling~\cite{hu2024codefilling}, and GenComm~\cite{zhou2025gencomm}. Here, E2E denotes an end-to-end baseline that jointly trains all parameters, including the ego encoder, auxiliary encoders, fusion module, and detection head. To assess generality across fusion architectures, we evaluate each method across various fusion backbones, including AttnFuse~\cite{xu2022opv2v}, CoBEVT~\cite{xu2022cobevt}, V2X-ViT~\cite{xu2022v2xvit}, and ParCon~\cite{bae2024parcon}. Detailed descriptions of the dataset, metrics, and implementation details are provided in Appendix~\ref{app:subsec:dataset}, \ref{app:subsec:metric}, and \ref{app:subsec:training_details}, respectively.



\paragraph{Experiment Design.}
\label{subsubsec:experiment_design}
We define the ego and auxiliary agent configurations separately. The ego agent is fixed to using a 128-beam LiDAR and a PointPillar backbone with voxel size $[0.4, 0.4, 30]$ along the $x$, $y$, and $z$ axes, respectively. For auxiliary agents, we consider two LiDAR specifications, 128-beam and 64-beam LiDARs, combined with four encoder configurations: two voxel-resolution variants of PointPillar~\cite{lang2019pointpillars} and two voxel-resolution variants of SECOND~\cite{yan2018second}, as summarized in Appendix~\ref{app:subsec:encoder_config}. This yields eight auxiliary agent configurations.

We define three evaluation protocols, where \EGO\ denotes the ego agent configuration (PP4 with a 128-beam LiDAR), \AUXONE\ denotes a \emph{seen} auxiliary agent configuration, and \AUXTWO\ denotes an \emph{unseen} auxiliary agent configuration:

\begin{itemize}[itemsep=0pt, topsep=0pt]
    \item[(1)] \textbf{Seen-pair evaluation.}
    The model is trained and evaluated on the same (\EGO\textbf{+}\AUXONE) pair.

    \item[(2)] \textbf{Zero-shot unseen evaluation.}
    The model is trained on (\EGO\textbf{+}\AUXONE) but evaluated on (\EGO\textbf{+}\AUXTWO) without additional training.

    \item[(3)] \textbf{Unseen adaptation evaluation.}
    The model is further fine-tuned on (\EGO\textbf{+}\AUXONE\textbf{+}\AUXTWO) and evaluated on (\EGO\textbf{+}\AUXTWO).
    Results are reported in Appendix~\ref{app:subsec:unseen_adaptation_evaluation}.
\end{itemize}
Protocol-specific application details are provided in Appendix~\ref{app:subsec:protocol-specific_evaluation_details}.

\subsection{Quantitative Analysis}
\label{subsec:quant_analysis}

\begin{figure}[t!]
\centering
\includegraphics[width=0.95\columnwidth]{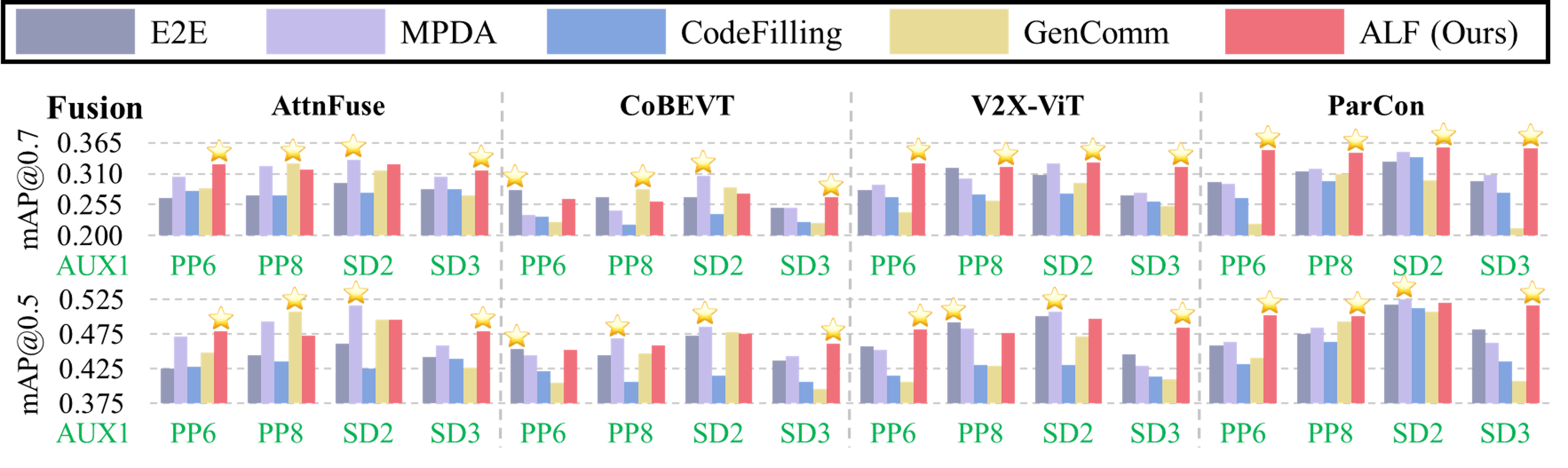}
\vspace{-8pt}
\caption*{(a) Grouped bar chart for seen-pair evaluation}
\vspace{3pt}
\begin{minipage}{0.95\columnwidth}
\centering
\footnotesize
\renewcommand{\arraystretch}{0.5}
\begin{tabular}{@{}c|ccccc@{}}
\toprule
Method & E2E & MPDA & CodeFilling & GenComm & \textbf{ALF (Ours)} \\
\midrule
\# Best@0.7 & 1 & 2 & 0 & 2 & \textbf{11} \\
\# Best@0.5 & 2 & 5 & 0 & 1 & \textbf{8} \\
\midrule
\begin{tabular}[c]{@{}c@{}}Bandwidth@10Hz\\ (Mbps) $\downarrow$\end{tabular}
& 503--1845 
& 503--1845 
& 1.475--2.703 
& 3.932--7.209 
& \textbf{0.0096} \\
\bottomrule
\end{tabular}
\end{minipage}
\vspace{-4pt}
\caption*{(b) Summary of the number of best-performing cases in terms of mAP@0.7 and mAP@0.5, together with bandwidth. Bandwidth is reported as a range because it varies across PP6, PP8, SD2, and SD3.}
\vspace{-4pt}
\caption{Comparison of detection accuracy under seen-pair evaluation. In this setting, both \EGO\ and \AUXONE\ use 128-beam LiDARs. Full results are provided in Appendix~\ref{app:subsec:seen_pair_evaluation}.}
\label{fig:sec4_seen_pair_evaluation}
\end{figure}

\paragraph{Seen-Pair Evaluation.}
Fig.~\ref{fig:sec4_seen_pair_evaluation} reports the results of the baseline methods under the seen-pair evaluation protocol and ALF for comparison. Despite its adaptation-free design, ALF remains competitive with SOTA methods that require pair-specific fine-tuning to handle heterogeneity. As shown in Fig.~\ref{fig:sec4_seen_pair_evaluation} (b), ALF achieves the highest mAP@0.5 and mAP@0.7 in 8 and 11 out of 16 cases, respectively. These results are particularly significant considering ALF communicates sub-1KB box-level messages, whereas feature-based SOTAs rely on dense, high-bandwidth representations.

\paragraph{Communication Bandwidth.}
The last row of Fig.~\ref{fig:sec4_seen_pair_evaluation}(b) reports the payload bitrate of each method at 10\,Hz. The transmitted message size of the baselines varies with the feature size determined by each agent configuration, leading to variable and often high communication costs. In contrast, ALF achieves a fixed and minimal bitrate because its message size is determined only by the fixed box schema and $K_{\max}$, not by the auxiliary feature size. Each message contains $K_{\max}=20$ boxes, with six INT8-quantized values per box, resulting in a fixed payload of $6 \times 20 \times 8 = 960$ bits, i.e., 120 bytes per frame. At a $10$\,Hz operating frequency, this corresponds to a bandwidth of only $9.6$\,kbps ($0.0096$\,Mbps), representing several orders of magnitude reduction compared to feature-based collaboration.
Although CodeFilling and GenComm transmit compact intermediate-derived messages, their communication cost remains tied to the BEV feature grid. 
In contrast, ALF communicates through fixed-size box-level messages, making its payload independent of the auxiliary feature-map resolution.

\begin{figure}[t!]
    \centering
    \includegraphics[width=0.99\linewidth]{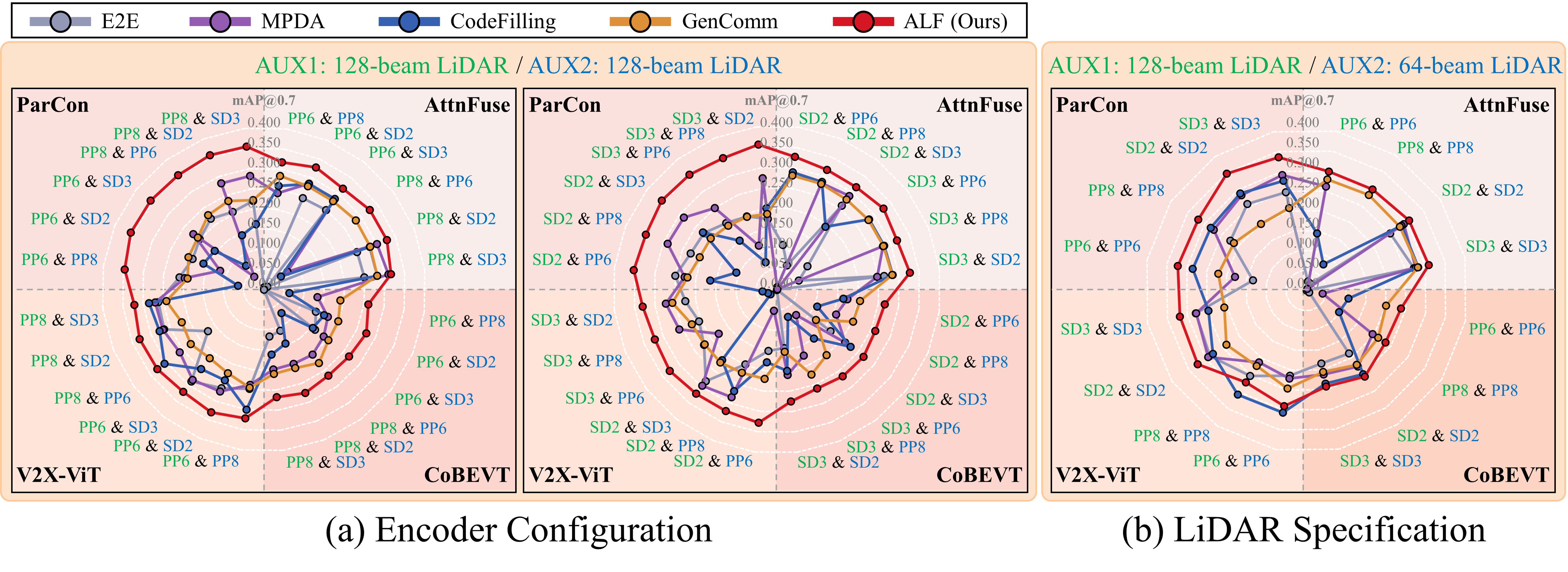}
    \vspace{-10pt}
    \caption{Comparison of detection accuracy under \textbf{zero-shot unseen evaluation} with two sources of heterogeneity: (a) encoder configuration and (b) sensor specification. In (a), both \EGO\ and \AUXONE\ use 128-beam LiDARs. In (b), \EGO\ uses a 128-beam LiDAR, while \AUXONE\ uses a 64-beam LiDAR. Full results are provided in Appendix~\ref{app:subsec:zero-shot_unseen_evaluation}.}
    \label{fig:sec4_zeroshot_evaluation}
    \vspace{-10pt}
\end{figure}

\paragraph{Zero-Shot Unseen Evaluation.}
Fig.~\ref{fig:sec4_zeroshot_evaluation} reports zero-shot unseen evaluation results for ALF and the baselines under two sources of heterogeneity: encoder configuration and LiDAR specification. 
As shown in Fig.~\ref{fig:sec4_zeroshot_evaluation}(a), ALF consistently outperforms all baselines under encoder-configuration heterogeneity. 
Fig.~\ref{fig:sec4_zeroshot_evaluation}(b) further shows that ALF achieves the best performance in most LiDAR-specification heterogeneity cases. 
These results demonstrate ALF's ability to handle unseen auxiliary configurations without additional adaptation.

Moreover, ALF maintains stable performance across both heterogeneity types and the individual unseen cases within each type. By contrast, E2E, MPDA, and CodeFilling show failure-prone cases, with mAP@0.7 dropping below 0.1. These failures indicate that configuration-specific baselines can be risky in open-world deployment. Although GenComm avoids such a dramatic drop, its performance is sensitive to the fusion method; under V2X-ViT and ParCon, it remains substantially below ALF.

\begin{figure}[t!]
    \centering
    \includegraphics[width=0.99\linewidth]{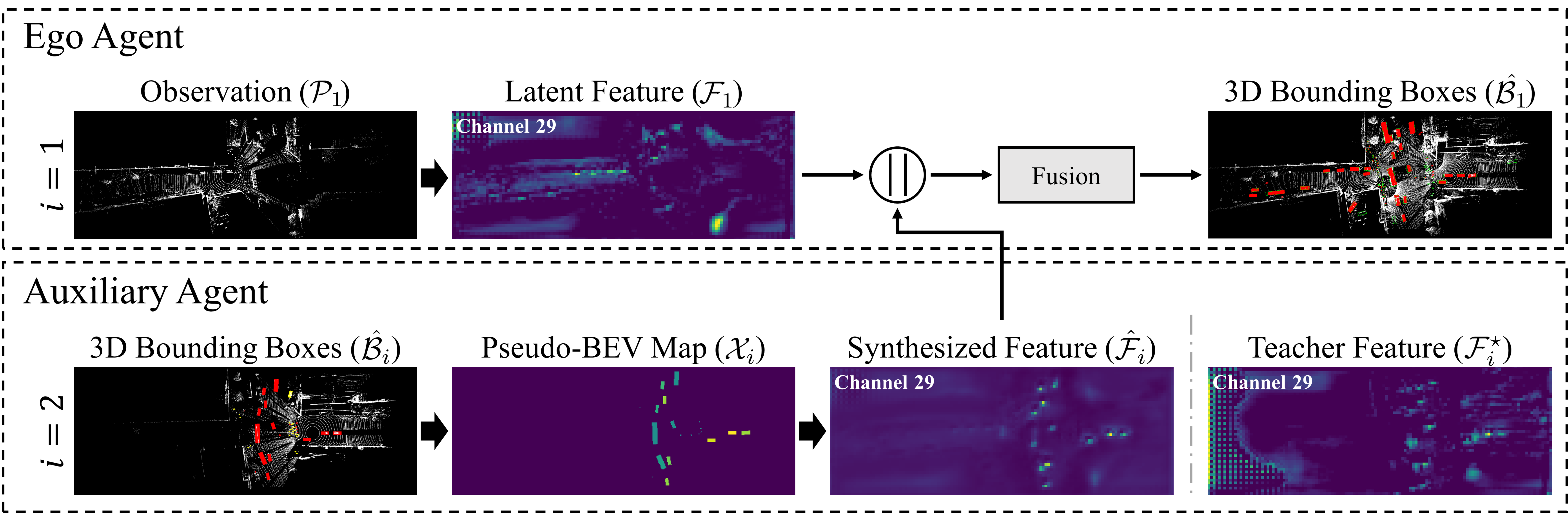}
    \caption{Visualization results of ALF using AttFuse. The 3D bounding boxes shown in red, green and yellow denote the predicted, ground-truth, and filtered out bounding boxes, respectively.}
    \label{fig:sec4_qualitative_analysis}
    \vspace{-13pt}
\end{figure}

\subsection{Qualitative Analysis}
\label{subsec:quali_analysis}
Fig.~\ref{fig:sec4_qualitative_analysis} visualizes the pseudo-BEV map $\mathcal{X}_i$, the synthesized feature $\hat{\mathcal{F}}_i$, the teacher feature $\mathcal{F}^{\star}_i$, and the ego-agent’s feature $\mathcal{F}_1$. We also provide the single-agent detection results $\{\hat{\mathcal{B}}_i\}_{i\in \mathcal{A}_{\mathrm{aux}}}$ and the final collaborated ego-centric detection $\hat{\mathcal{B}}_1$.
Using EFS, the ego agent transforms object-centric cues from the pseudo-BEV map into ego-compatible latent features by leveraging the geometric and semantic context of $\mathcal{F}_1$. 
Consequently, the synthesized feature $\hat{\mathcal{F}}_i$ integrates the auxiliary object-centric cue with ego scene context, yielding a fusion-ready representation in the ego latent space.

\subsection{Ablation Study}
\label{subsec:abla_study}
All ablation studies are conducted under the homogeneous encoder setting PP4 with a 128-beam LiDAR. Fig.~\ref{fig:sec4_abl_data_size} shows detection performance under different communication bandwidth settings. The blue curve compares quantization precisions (INT4, INT8, INT16, and FP32). Quantizing the transmitted data to INT4 causes a clear drop in mAP@0.7
, indicating substantial information loss. By contrast, increasing precision beyond INT8 yields only marginal gains.
This suggests that INT8 provides the best trade-off between communication efficiency and detection performance.

The yellow curve shows the effect of varying the maximum number of transmitted bounding boxes $K_{\max}$. When $K_{\max}$ is 0, corresponding to no fusion, the mAP@0.7 is 0.295. Increasing the limit to 20 improves the performance to 0.333. However, transmitting more than 20 boxes degrades performance.
This suggests that additional low-confidence boxes provide little information and may hurt detection.

Tab.~\ref{tab:sec4_abl_efs} presents an ablation study of EFS. Directly using the pseudo-BEV map without EFS gives the worst performance, indicating that the object-centric cue cannot be consumed by the ego latent space as-is. 
The full EFS achieves the best performance, showing that OCE, EIM, and ELR are complementary for synthesizing ego-compatible latent features.

\begin{table}[t!]
\centering
\begin{tabular}{@{}m{0.5\columnwidth}@{\hspace{0.02\columnwidth}}m{0.46\columnwidth}@{}}
\begin{minipage}[t]{\linewidth}
\vspace{0pt}
\centering
\includegraphics[width=0.92\linewidth]{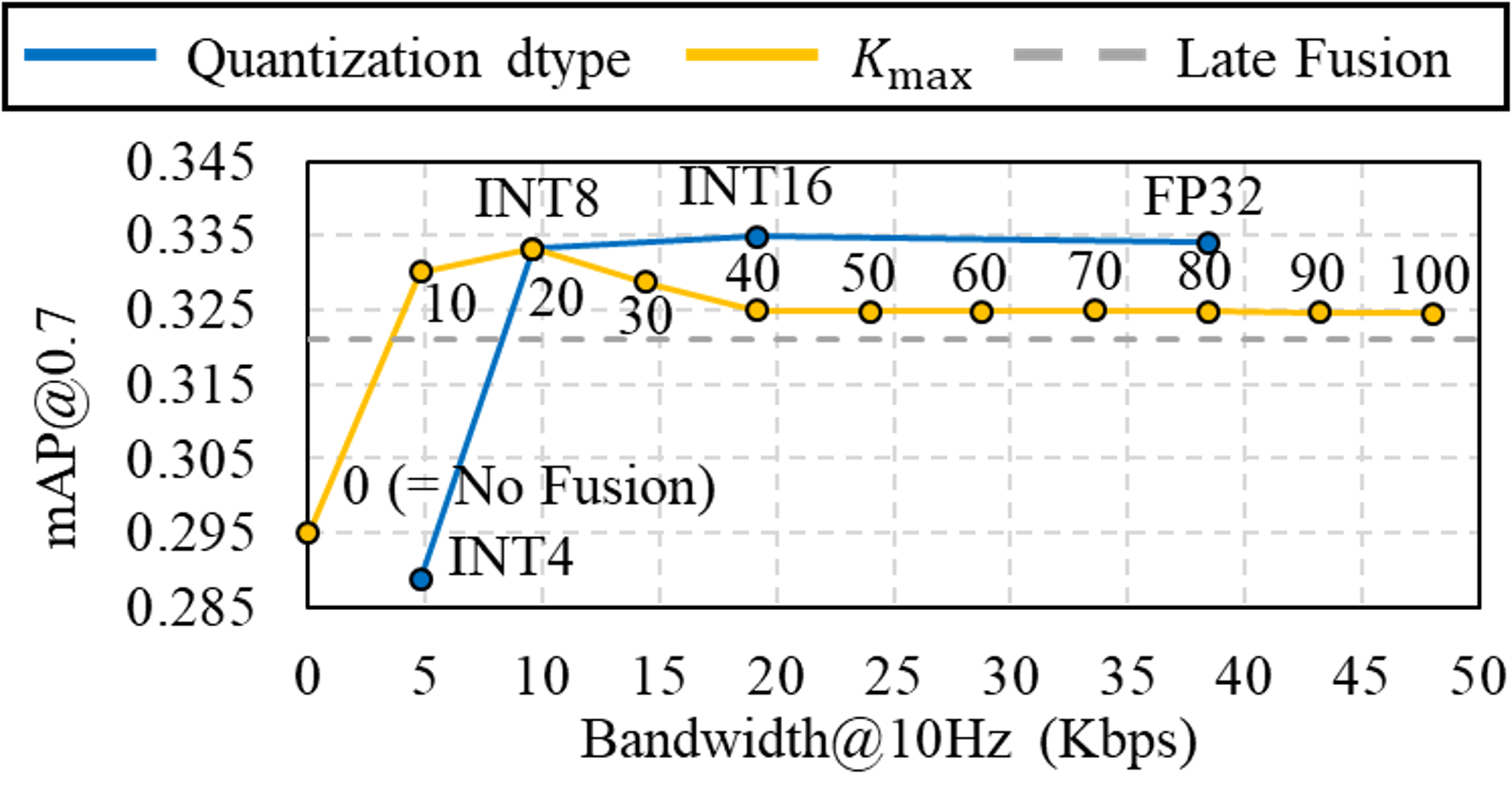}
\captionof{figure}{Ablation study on quantization schemes, and maximum number of transmitted boxes $K_{\max}$.}
\label{fig:sec4_abl_data_size}
\end{minipage}
&
\begin{minipage}[t]{\linewidth}
\vspace{0pt}
\centering
\footnotesize
\captionof{table}{Ablation study on EFS using different combinations of its submodules.}
\label{tab:sec4_abl_efs}
\renewcommand{\arraystretch}{1.2}
\begin{tabular}{@{}ccc|cc@{}}
\toprule
OCE & EIM & ELR & mAP@0.5 & mAP@0.7 \\ \midrule
 &  &  & 0.452 & 0.298 \\
$\checkmark$ &  &  & 0.470 & 0.305 \\
$\checkmark$ & $\checkmark$ &  & 0.483 & 0.313 \\
$\checkmark$ &  & $\checkmark$ & 0.487 & 0.318 \\
$\checkmark$ & $\checkmark$ & $\checkmark$ & \strokeBold{0.485} & \strokeBold{0.333} \\
\bottomrule
\end{tabular}
\end{minipage}
\end{tabular}
\vspace{-20pt}
\end{table}

\section{Conclusion}
\label{sec:conclusion}
We study configuration-agnostic heterogeneity in cooperative 3D object detection, where auxiliary agents may have unseen sensor specifications or encoder configurations. In open-world deployment, such auxiliary agents may appear after training, and the ego agent must collaborate with them without further adaptation.
To address this challenge, we propose ALF, a late-to-intermediate fusion framework that converts lightweight box-level messages into pseudo-BEV map and synthesizes ego-compatible auxiliary features using ego-scene context. 
Experiments on V2X-Real show that ALF handles unseen auxiliary configurations under zero-shot evaluation while maintaining low communication cost. These results suggest that ego-side synthesis from compact box-level messages provides a practical and scalable interface for heterogeneous collaborative perception.

\textbf{Limitations and Future Work.}
To isolate the effect of configuration heterogeneity, we do not consider additional open-world deployment factors such as sensor noise, communication noise, localization error, or latency. 
Future work will extend ALF to such realistic deployment settings.
\bibliographystyle{abbrvnat}
\bibliography{alfbib}

\appendix

\section{Method Details}
\label{app:sec:detailed_method}
\subsection{Quantization Details}
\label{app:subsec:quantization_details}
For each scalar field $v$, we apply $b$-bit zero-point quantization:
\begin{equation}
\tilde{v} =
\mathrm{clip}\!\left(\left\lfloor\frac{v}{s_v}\right\rceil + z_v,\ q_{\min}, q_{\max}\right),\quad
s_v = \frac{v_{\max}-v_{\min}}{q_{\max}-q_{\min}},\quad
z_v = \left\lfloor q_{\min}-\frac{v_{\min}}{s_v}\right\rceil,
\end{equation}
where $q_{\min}=0$ and $q_{\max}=2^b-1$.

\subsection{Details of the Box-to-BEV Rasterizer}
\label{app:subsec:b2br_details}

\paragraph{Dequantization and Ego-Frame Transformation.}
Upon reception, the ego agent dequantizes each field via
\begin{equation}
v^{\mathrm{dq}} = s_v(\tilde{v}-z_v),
\end{equation}
and transforms $(x^{\mathrm{dq}}, y^{\mathrm{dq}}, \psi^{\mathrm{dq}})$ to the ego frame using $T_{i\rightarrow 1}$.

\paragraph{Rasterization Details.}
B2BR converts the processed box message into a single-channel pseudo-BEV map by rasterizing box footprints on the BEV grid. Each box contributes a confidence-weighted occupancy pattern, and the final map stores the occupied cells after rasterization.

The grid resolution is determined by the ego agent's BEV detection range $[x_{\min}, x_{\max}] \times [y_{\min}, y_{\max}]$ and the voxel size $(v_x, v_y)$:
\[
H_{\mathrm{bev}}=\left\lfloor\frac{x_{\max}-x_{\min}}{v_x}\right\rfloor,
\qquad
W_{\mathrm{bev}}=\left\lfloor\frac{y_{\max}-y_{\min}}{v_y}\right\rfloor.
\]

For a transformed box parameterized by $(x,y,\ell,w,\psi)$ on the BEV plane, we test whether each grid cell center lies inside the rotated box footprint. Let $(x_{g,u}, y_{g,v})$ denote the center of grid cell $(u,v)$. We first translate the cell center to the box center and then rotate it into the box-local coordinate frame:
\begin{align}
\delta x &= x_{g,u} - x, \qquad \delta y = y_{g,v} - y, \\
\begin{bmatrix} x' \\ y' \end{bmatrix}
&=
\begin{bmatrix}
\cos\psi & \sin\psi \\
-\sin\psi & \cos\psi
\end{bmatrix}
\begin{bmatrix} \delta x \\ \delta y \end{bmatrix}.
\end{align}
The cell is marked as occupied if
\begin{equation}
|x'|\le \frac{\ell}{2}, \qquad |y'|\le \frac{w}{2}.
\end{equation}

If multiple boxes overlap in the same cell, we assign the maximum confidence score among them. This produces a sparse pseudo-BEV map that preserves the spatial footprint of auxiliary detections while remaining independent of the auxiliary encoder configuration.

\subsection{Detailed Architecture of the Ego-compatible Feature Synthesizer}
\label{app:subsec:EFS_Architecture}

\paragraph{Basic operators.}
We denote a 2D convolution with kernel size $k$ and stride $s$ by $\mathrm{Conv}_{k,s}$, batch normalization by $\mathrm{BN}$, and the Sigmoid Linear Unit by $\mathrm{SiLU}$. 
We define
\begin{align}
\mathrm{CBA}_{k,s}(\cdot)
&\triangleq
\mathrm{Conv}_{k,s}\rightarrow\mathrm{BN}\rightarrow\mathrm{SiLU}, \\
\mathrm{Bilinear}_{\times m}(\cdot)
&\triangleq 
\text{bilinear interpolation by a factor of }m.
\end{align}
We further use a residual block
\begin{equation}
\mathrm{RB}(\mathbf{x})
\triangleq
\mathbf{x}
+
\mathrm{BN}\!\left(
\mathrm{Conv}_{3,1}\!\left(
\mathrm{SiLU}\!\left(
\mathrm{BN}\!\left(
\mathrm{Conv}_{3,1}(\mathbf{x})
\right)\right)\right)\right).
\end{equation}

\paragraph{Object-centric Cue Encoder (OCE).}
The operator $\mathrm{Expand}(\cdot)$ performs channel-wise expansion by replication:
\begin{equation}
\mathrm{Expand}(\mathcal{X}_i)
=
\underbrace{[\mathcal{X}_i,\dots,\mathcal{X}_i]}_{\text{$C_0$ times}}
\in \mathbb{R}^{H_{\mathrm{bev}}\times W_{\mathrm{bev}}\times C_0},
\end{equation}
where $C_0=64$ in our implementation.
This operation introduces no learnable parameters and simply duplicates the single-channel input along the channel dimension.

The encoder $\mathrm{Enc}_{L}(\cdot)$ consists of $L$ stages:
\begin{equation}
\mathcal{O}_i^{(\ell)}
=
\mathrm{CBA}_{3,1}\!\left(
\mathrm{CBA}_{3,2}\!\left(\mathcal{O}_i^{(\ell-1)}\right)
\right),
\quad \ell=1,\dots,L,
\end{equation}
with $\mathcal{O}_i^{(0)}=\mathrm{Expand}(\mathcal{X}_i)$.
Here, $\mathrm{CBA}_{3,2}$ reduces spatial resolution by a factor of $2$, while $\mathrm{CBA}_{3,1}$ refines features at the same scale. Across the $L$ stages, the spatial resolution is reduced by a factor of $2^L$, while the channel dimension is progressively increased to reach the output dimension of $\mathcal{O}_i$.

\paragraph{Ego-context Injection Module (EIM).}
The ego feature is first spatially aligned by average pooling:
\begin{equation}
\bar{\mathcal{F}}_1
=
\mathrm{AvgPool}(\mathcal{F}_1),
\end{equation}
where $\mathrm{AvgPool}(\cdot)$ denotes adaptive average pooling whose output spatial resolution matches that of the feature to which it is added.

The downsampling operators are defined as
\begin{equation}
\mathrm{Down}_{k}(\mathbf{x})
=
\mathrm{CBA}_{3,2}(\mathbf{x}),
\quad k\in\{1,2\}.
\end{equation}

The upsampling operators restore resolution via bilinear interpolation followed by refinement:
\begin{equation}
\mathrm{Up}_{k}(\mathbf{x})
=
\mathrm{CBA}_{3,1}\!\left(
\mathrm{Bilinear}_{\times s_k}(\mathbf{x})
\right),
\end{equation}
where $s_1=2$ and $s_2=4$.

Thus, the two ego-enriched features are instantiated as
\begin{align}
\mathcal{Z}_i'^{(1)}
&=
\mathrm{Up}_{1}\!\left(
\mathrm{Down}_{1}\!\left(
\mathcal{O}_i+\mathrm{AvgPool}(\mathcal{F}_1)
\right)
\right), \\
\mathcal{Z}_i'^{(2)}
&=
\mathrm{Up}_{2}\!\left(
\mathrm{Down}_{2}\!\left(
\mathrm{Down}_{1}(\mathcal{O}_i)+\mathrm{AvgPool}(\mathcal{F}_1)
\right)
\right).
\end{align}

\paragraph{Ego-compatible Latent Refiner (ELR).}
The projection operator is defined as
\begin{equation}
\mathrm{Proj}(\mathbf{x})
=
\mathrm{CBA}_{3,1}\!\big(
\mathrm{CBA}_{3,1}(\mathbf{x})
\big),
\end{equation}
which maps the concatenated feature to channel dimension $C$.

The refinement operator is defined as
\begin{equation}
\mathrm{ResBlock}_{3}(\mathbf{x})
=
(\mathrm{RB}\circ\mathrm{RB}\circ\mathrm{RB})(\mathbf{x}),
\end{equation}
which applies three residual blocks while preserving spatial resolution.


\subsection{Teacher Feature Construction}
\label{app:subsec:teacher_feature}
For each auxiliary agent $i$, we define a teacher feature $\mathcal{F}_i^{\star}$ as the latent feature produced by the frozen ego encoder when the auxiliary observation is expressed in the ego coordinate frame. Specifically, let $\mathcal{P}_i$ denote the auxiliary observation and let $T_{i\rightarrow1}$ denote the relative pose from agent $i$ to the ego agent. We first transform $\mathcal{P}_i$ into the ego frame and denote the transformed observation by $\tilde{\mathcal{P}}_{i\rightarrow1}$. The teacher feature is then obtained by encoding $\tilde{\mathcal{P}}_{i\rightarrow1}$ with the frozen ego encoder.

This construction differs from using the auxiliary agent's native latent feature. Instead, it defines the supervisory target directly in the latent space expected by the ego fusion and detection stack. As a result, the alignment objective encourages the synthesized feature to be compatible with the ego latent domain while preserving the foreground structure induced by the auxiliary observation.

\subsection{Pseudo-BEV-Based Mask Construction for Region-Aware Alignment}
\label{app:subsec:pseudo_bev_mask}

We provide implementation details for constructing masks derived from the pseudo-BEV map. Let $\mathcal{X}_i \in \mathbb{R}^{H_{\mathrm{bev}} \times W_{\mathrm{bev}} \times 1}$ denote the pseudo-BEV map for auxiliary agent $i$. We downsample $\mathcal{X}_i$ to the target spatial resolution $(H, W)$ of the ego latent feature using max pooling:
\begin{equation}
\bar{\mathcal{X}}_i = \mathrm{MaxPool}(\mathcal{X}_i; H, W).
\end{equation}

We first obtain a foreground occupancy mask by thresholding the downsampled map:
\begin{equation}
\mathbf{\Omega}_i^{\mathrm{fg}} = \mathbf{1}[\bar{\mathcal{X}}_i > \tau].
\end{equation}
Because this mask is often too sparse, we further expand it using morphological dilation to obtain the object-region mask used in the main loss:
\begin{equation}
\mathbf{\Omega}_i^{\mathrm{obj}} = \mathrm{Dilate}(\mathbf{\Omega}_i^{\mathrm{fg}}; d_r),
\end{equation}
where $d_r$ denotes the dilation radius, and the corresponding kernel size is $2d_r+1$. The background mask is then defined as the complement of the object-region mask:
\begin{equation}
\mathbf{\Omega}_i^{\mathrm{bg}} = \mathbf{1} - \mathbf{\Omega}_i^{\mathrm{obj}}.
\end{equation}

This construction allows the alignment loss to emphasize not only object-occupied regions but also their surrounding context. In all experiments, we set $\tau = 0$, use max pooling for spatial downsampling, and set the dilation radius to $d_r = 2$.

\section{Experiments details}
\label{app:sec:experiments_detailse}

\subsection{Dataset} 
\label{app:subsec:dataset}
We evaluate all methods on V2X-Real, a large-scale real-world V2X cooperative perception dataset collected with two connected automated vehicles and two smart infrastructure units, each equipped with LiDAR and multi-view cameras. It contains 33K LiDAR frames and 171K camera images with over 1.2M annotated 3D bounding boxes across 10 categories, and provides four subsets (vehicle-centric, infrastructure-centric, V2V, and I2I) to evaluate diverse collaboration modes.

\subsection{Metrics}
\label{app:subsec:metric}
We group the 10 annotated categories into three superclasses: car, pedestrian, and truck.
For each IoU threshold, we compute the Average Precision (AP) for each superclass and report the mean Average Precision (mAP) as 
$\mathrm{mAP}=\frac{1}{3}\left(\mathrm{AP}_{\text{car}}+\mathrm{AP}_{\text{ped}}+\mathrm{AP}_{\text{truck}}\right)$.
We denote mAP@0.5 and mAP@0.7 for thresholds of 0.5 and 0.7, respectively.

\subsection{Encoder Configurations}
\label{app:subsec:encoder_config}
We define five encoder configurations based on backbone architecture and voxel size, as summarized in Tab.~\ref{tab:app_encoder_configuration}. PP4 is used only for the ego agent, while the remaining four encoder configurations are used for auxiliary agents.

\begin{table}[t!]
\caption{Encoder Configurations. Performance is evaluated using single-agent detectors without fusion (no fusion).}
\label{tab:app_encoder_configuration}
\centering
\scriptsize
\begin{tabular}{c|c|c|c|c|c}
\toprule
\begin{tabular}[c]{@{}c@{}}Backbone\\ Architecture\end{tabular} & \begin{tabular}[c]{@{}c@{}}Encoder\\ Configuration\end{tabular} & \begin{tabular}[c]{@{}c@{}}Voxel Size\\ ($v_x, v_y, v_z$)\end{tabular} & \begin{tabular}[c]{@{}c@{}}2D / 3D\\ CNN Layers\end{tabular} & \begin{tabular}[c]{@{}c@{}}Detection Range \\ $\left([x_{\min}, x_{\max}] \times [y_{\min}, y_{\max}]\right)$ \end{tabular} & \begin{tabular}[c]{@{}c@{}}Feature Size\\ ($H \times W \times C$)\end{tabular}\\ \midrule
\multirow{3}{*}{PointPillar} & PP4 & 0.4, 0.4, 30 & 21 / 0 & [-102.4, 102.4] $\times$ [-38.4, 38.4] & $48 \times 128 \times 256$\\
 & PP6 & 0.6, 0.6, 30 & 21 / 0 & [-105.6, 105.6] $\times$ [-38.4, 38.4] & $64\times 176\times256$ \\
 & PP8 & 0.8, 0.8, 30 & 21 / 0 & [-102.4, 102.4] $\times$ [-38.4, 38.4] & $48 \times 128 \times 256$\\ \midrule
\multirow{2}{*}{SECOND} & SD2 & 0.2, 0.2, 0.1 & 12 / 12 & [-102.4, 102.4] $\times$ [-38.4, 38.4] & $48 \times 128 \times 512$\\
 & SD3 & 0.3, 0.3, 0.1 & 12 / 12 & [-105.6, 105.6] $\times$ [-38.4, 38.4] & $64 \times 176 \times 512$\\ \bottomrule
\end{tabular}
\end{table}


\subsection{Training Details}
\label{app:subsec:training_details}
Our training procedure consists of encoder pretraining followed by separate fine-tuning protocols for ALF and baselines. All models are trained and evaluated on an RTX 4090 GPU.

\paragraph{Encoder Pretraining.}
For each agent configuration, we pretrain three types of checkpoints: (1) single-agent detectors, which are used for late fusion; (2) homogeneous intermediate-fusion models, where all agents share the same encoder configuration; and (3) GenComm~\cite{zhou2025gencomm} stage-1 models.

For the standalone single-agent detectors used in late fusion, we train each model for 200 epochs using AdamW~\cite{loshchilov2017adamw} with an initial learning rate of $3{\times}10^{-4}$ and weight decay of $1{\times}10^{-2}$. We adopt cosine annealing with warm-up. The warm-up lasts for the first 40 epochs, starts from a learning rate of $2{\times}10^{-4}$, and the minimum learning rate is set to $2{\times}10^{-5}$.

For the homogeneous intermediate-fusion models and GenComm stage-1 models, we use the same optimizer and scheduler, but train for 40 epochs with warm-up during the first 10 epochs.

\paragraph{ALF Fine-tuning.}
For ALF, we fine-tune the EFS parameters only once, while keeping all other components frozen, using the initialization described in Sec.~\ref{subsec:training_and_deployment}. The maximum number of transmitted bounding boxes $K_{\max}$ is set to 60 during training and 20 during inference. All transmitted data are quantized from FP32 to INT8. The loss weights are set to $\alpha_{\mathrm{cls}}=1.0$, $\alpha_{\mathrm{reg}}=2.0$, $\alpha_{\mathrm{obj}}=1.0$, $\alpha_{\mathrm{bg}}=0.5$, $\alpha_{\mathrm{det}}=1.0$, and $\alpha_{\mathrm{align}}=1.0$.

\paragraph{Baseline-specific Fine-tuning.}
During baseline fine-tuning, E2E, MPDA, and CodeFilling initialize the ego and auxiliary branches from the homogeneous intermediate-fusion checkpoints of their corresponding agent configurations, while GenComm initializes from the corresponding stage-1 checkpoints. Each baseline is fine-tuned for 10 epochs using Adam~\cite{kingma2015adam} with an initial learning rate of $1{\times}10^{-3}$, $\epsilon{=}1{\times}10^{-10}$, and weight decay of $1{\times}10^{-4}$. We use a multi-step decay schedule with decay factor $\gamma{=}0.1$ and a milestone at epoch 5.

During baseline-specific fine-tuning, we follow the freezing strategy of each method. E2E is trained end-to-end without freezing any layers. MPDA and CodeFilling keep the ego and auxiliary encoders frozen, while GenComm updates only the DME and freezes all remaining layers.

Unlike ALF, which fine-tunes the EFS only once and reuses it across agent configurations, the baselines must be fine-tuned separately for each agent configuration.

\subsection{Protocol-Specific Evaluation Details}
\label{app:subsec:protocol-specific_evaluation_details}
Below, we describe how the evaluated models are defined and applied under each protocol.
\begin{itemize}
    \item \textbf{Baselines.}
    For the baselines, protocols (1)--(3) in Sec.~\ref{subsubsec:experiment_design} correspond to protocol-dependent training and evaluation under different auxiliary agent configurations. In the zero-shot unseen setting, to apply a model trained with \AUXONE\ to \AUXTWO, we align the output representation of \AUXTWO\ to the auxiliary feature shape used with \AUXONE\ (Tab.~\ref{tab:app_encoder_configuration}). Specifically, we match the channel dimension using a frozen $1{\times}1$ convolution with orthogonally initialized weights, and resample the BEV spatial resolution with bilinear interpolation.

    \item \textbf{ALF.}
    ALF does not require protocol-specific training or feature-shape alignment when the auxiliary encoder changes. Therefore, its results are determined only by the auxiliary agent configuration used at inference and are invariant to the protocol label.
\end{itemize}



\subsection{Full Results of Seen-Pair Evaluation}
\label{app:subsec:seen_pair_evaluation}
Tab.~\ref{tab:app_seen-pair_eval} provides the full quantitative results corresponding to Fig.~\ref{fig:sec4_seen_pair_evaluation}.

\begin{table}[h!]
\caption{Seen-pair Evaluation (\EGO+\AUXONE). The best performance is highlighted in bold, and the second-best is underlined. In this setting, both \EGO\ and \AUXONE\ use 128-beam LiDARs.}
\label{tab:app_seen-pair_eval}
\centering
\footnotesize
\begin{tabular}{c|c|ccccc@{}}
\toprule
 \multirow{2}{*}{\begin{tabular}[c]{@{}c@{}}Fusion \\ Module\end{tabular}} & \multirow{2}{*}{\AUXONE} & \multicolumn{1}{c|}{E2E} & \multicolumn{1}{c|}{MPDA} & \multicolumn{1}{c|}{CodeFilling} & \multicolumn{1}{c|}{GenComm} & AFL (Ours) \\
 &  & \multicolumn{1}{c|}{mAP@0.5/0.7} & \multicolumn{1}{c|}{mAP@0.5/0.7} & \multicolumn{1}{c|}{mAP@0.5/0.7} & \multicolumn{1}{c|}{mAP@0.5/0.7} & mAP@0.5/0.7 \\ \midrule
 & PP6 & \multicolumn{1}{c|}{0.425 /   0.266} & \multicolumn{1}{c|}{\uline{0.471} / \uline{0.305}} & \multicolumn{1}{c|}{0.428 / 0.279} & \multicolumn{1}{c|}{0.448 / 0.284} & \strokeBold{0.479} / \strokeBold{0.327} \\
\multirow{2}{*}{AttnFuse} & PP8 & \multicolumn{1}{c|}{0.444 / 0.272} & \multicolumn{1}{c|}{\uline{0.493} / \uline{0.323}} & \multicolumn{1}{c|}{0.435 / 0.271} & \multicolumn{1}{c|}{\strokeBold{0.507} / \strokeBold{0.329}} & 0.472 / 0.317 \\
 & SD2 & \multicolumn{1}{c|}{0.461 / 0.293} & \multicolumn{1}{c|}{\strokeBold{0.516} / \strokeBold{0.335}} & \multicolumn{1}{c|}{0.425 / 0.276} & \multicolumn{1}{c|}{\uline{0.496} / 0.316} & 0.496 / \uline{0.327} \\
 & SD3 & \multicolumn{1}{c|}{0.442 / 0.282} & \multicolumn{1}{c|}{\uline{0.459} / \uline{0.304}} & \multicolumn{1}{c|}{0.440 / 0.282} & \multicolumn{1}{c|}{0.427 / 0.272} & \strokeBold{0.479} / \strokeBold{0.316} \\ \midrule
 & PP6 & \multicolumn{1}{c|}{\strokeBold{0.453} / \strokeBold{0.281}} & \multicolumn{1}{c|}{0.445 / 0.236} & \multicolumn{1}{c|}{0.421 / 0.234} & \multicolumn{1}{c|}{0.405 / 0.224} & \uline{0.452} / \uline{0.266} \\
\multirow{2}{*}{CoBEVT} & PP8 & \multicolumn{1}{c|}{0.445 / \uline{0.269}} & \multicolumn{1}{c|}{\strokeBold{0.469} / 0.245} & \multicolumn{1}{c|}{0.406 / 0.219} & \multicolumn{1}{c|}{0.447 / \strokeBold{0.282}} & \uline{0.458} / 0.261 \\
 & SD2 & \multicolumn{1}{c|}{0.473 / 0.268} & \multicolumn{1}{c|}{\strokeBold{0.486} / \strokeBold{0.306}} & \multicolumn{1}{c|}{0.415 / 0.238} & \multicolumn{1}{c|}{\uline{0.478} / \uline{0.286}} & 0.475 / 0.275 \\
 & SD3 & \multicolumn{1}{c|}{0.437 / 0.249} & \multicolumn{1}{c|}{\uline{0.443} / \uline{0.250}} & \multicolumn{1}{c|}{0.406 / 0.224} & \multicolumn{1}{c|}{0.396 / 0.223} & \strokeBold{0.461} / \strokeBold{0.268} \\ \midrule
 & PP6 & \multicolumn{1}{c|}{\uline{0.457} / 0.281} & \multicolumn{1}{c|}{0.452 / \uline{0.291}} & \multicolumn{1}{c|}{0.415 / 0.268} & \multicolumn{1}{c|}{0.406 / 0.241} & \strokeBold{0.482} / \strokeBold{0.328} \\
\multirow{2}{*}{V2X-ViT} & PP8 & \multicolumn{1}{c|}{\strokeBold{0.492} / \uline{0.320}} & \multicolumn{1}{c|}{\uline{0.483} / 0.301} & \multicolumn{1}{c|}{0.430 / 0.273} & \multicolumn{1}{c|}{0.429 / 0.262} & 0.477 / \strokeBold{0.322} \\
 & SD2 & \multicolumn{1}{c|}{\uline{0.501} / 0.307} & \multicolumn{1}{c|}{\strokeBold{0.507} / \uline{0.329}} & \multicolumn{1}{c|}{0.430 / 0.275} & \multicolumn{1}{c|}{0.471 / 0.293} & 0.498 / \strokeBold{0.330} \\
 & SD3 & \multicolumn{1}{c|}{\uline{0.446} / 0.271} & \multicolumn{1}{c|}{0.429 / \uline{0.276}} & \multicolumn{1}{c|}{0.414 / 0.261} & \multicolumn{1}{c|}{0.410 / 0.252} & \strokeBold{0.485} / \strokeBold{0.322} \\ \midrule
\multicolumn{1}{l|}{} & PP6 & \multicolumn{1}{c|}{0.459 / \uline{0.295}} & \multicolumn{1}{c|}{\uline{0.464} / 0.292} & \multicolumn{1}{c|}{0.432 / 0.267} & \multicolumn{1}{c|}{0.441 / 0.221} & \strokeBold{0.503} / \strokeBold{0.352} \\
\multicolumn{1}{c|}{\multirow{2}{*}{ParCon}} & PP8 & \multicolumn{1}{c|}{0.475 / 0.314} & \multicolumn{1}{c|}{0.484 / \uline{0.319}} & \multicolumn{1}{c|}{0.464 / 0.297} & \multicolumn{1}{c|}{\uline{0.493} / 0.309} & \strokeBold{0.501} / \strokeBold{0.347} \\
\multicolumn{1}{l|}{} & SD2 & \multicolumn{1}{c|}{0.517 / 0.331} & \multicolumn{1}{c|}{\strokeBold{0.525} / \uline{0.348}} & \multicolumn{1}{c|}{0.513 / 0.339} & \multicolumn{1}{c|}{0.508 / 0.299} & \uline{0.520} / \strokeBold{0.357} \\
\multicolumn{1}{l|}{} & SD3 & \multicolumn{1}{c|}{\uline{0.482} / 0.296} & \multicolumn{1}{c|}{0.462 / \uline{0.308}} & \multicolumn{1}{c|}{0.436 / 0.276} & \multicolumn{1}{c|}{0.407 / 0.213} & \strokeBold{0.516} / \strokeBold{0.355} \\ \bottomrule
\end{tabular}
\end{table}

\subsection{Full Results of Zero-Shot Unseen Evaluation}
\label{app:subsec:zero-shot_unseen_evaluation}
Tab.~\ref{tab:app_unseen-zero-shot_eval} presents the full quantitative results for the heterogeneous encoder configurations shown in Fig.~\ref{fig:sec4_zeroshot_evaluation}(a).

\begin{table}[h!]
\caption{Zero-shot unseen evaluation on (\EGO+\AUXTWO). Models trained/fine-tuned on (\EGO+\AUXONE) are directly evaluated on (\EGO+\AUXTWO) without additional training. To enable transfer from \AUXONE to \AUXTWO, the output representation of \AUXTWO\ is aligned to the auxiliary feature size used with \AUXONE. The best performance is highlighted in bold. In this setting, \EGO, \AUXONE, and \AUXTWO\ are all equipped with 128-beam LiDARs.}
\label{tab:app_unseen-zero-shot_eval}
\centering
\footnotesize
\begin{tabular}{@{}ccc|ccccc}
\toprule
Fusion & \multirow{2}{*}{\AUXONE} & \multirow{2}{*}{\AUXTWO} & E2E & MPDA & CodeFilling & GenComm & ALF (Ours) \\
Method &  &  & mAP@0.5/0.7 & mAP@0.5/0.7 & mAP@0.5/0.7 & mAP@0.5/0.7 & mAP@0.5/0.7 \\ \midrule
 &  & PP8 & 0.052 /   0.005 & 0.379 /   0.240 & 0.395 /   0.258 & 0.435 /   0.283 & \strokeBold{0.472} / \strokeBold{0.317} \\
 & PP6 & SD2 & 0.407 / 0.245 & 0.443 / 0.284 & 0.433 / 0.283 & 0.429 / 0.277 & \strokeBold{0.496} / \strokeBold{0.327} \\
 &  & SD3 & 0.406 / 0.249 & 0.444 / 0.283 & 0.435 / 0.284 & 0.429 / 0.277 & \strokeBold{0.479} / \strokeBold{0.316} \\ \cmidrule(l){2-8} 
 &  & PP6 & 0.079 / 0.011 & 0.103 / 0.073 & 0.079 / 0.053 & 0.437 / 0.284 & \strokeBold{0.479} / \strokeBold{0.327} \\
 & PP8 & SD2 & 0.404 / 0.248 & 0.451 / 0.301 & 0.444 / 0.282 & 0.435 / 0.283 & \strokeBold{0.496} / \strokeBold{0.327} \\
\multirow{2}{*}{AttnFuse} &  & SD3 & 0.399 / 0.250 & 0.457 / 0.310 & 0.444 / 0.282 & 0.434 / 0.282 & \strokeBold{0.479} / \strokeBold{0.316} \\ \cmidrule(l){2-8} 
 &  & PP6 & 0.257 / 0.110 & 0.014 / 0.006 & 0.446 / 0.289 & 0.428 / 0.281 & \strokeBold{0.479} / \strokeBold{0.327} \\
 & SD2 & PP8 & 0.182 / 0.064 & 0.444 / 0.282 & 0.446 / 0.285 & 0.426 / 0.280 & \strokeBold{0.472} / \strokeBold{0.317} \\
 &  & SD3 & 0.413 / 0.259 & 0.448 / 0.289 & 0.284 / 0.194 & 0.425 / 0.278 & \strokeBold{0.479} / \strokeBold{0.316} \\ \cmidrule(l){2-8} 
 &  & PP6 & 0.284 / 0.094 & 0.004 / 0.002 & 0.441 / 0.283 & 0.426 / 0.281 & \strokeBold{0.479} / \strokeBold{0.327} \\
 & SD3 & PP8 & 0.168 / 0.058 & 0.448 / 0.280 & 0.440 / 0.283 & 0.426 / 0.281 & \strokeBold{0.472} / \strokeBold{0.317} \\
 &  & SD2 & 0.409 / 0.264 & 0.374 / 0.247 & 0.444 / 0.285 & 0.428 / 0.284 & \strokeBold{0.496} / \strokeBold{0.327} \\ \midrule
 &  & PP8 & 0.008 / 0.001 & 0.277 / 0.134 & 0.129 / 0.064 & 0.338 / 0.191 & \strokeBold{0.458} / \strokeBold{0.261} \\
 & PP6 & SD2 & 0.266 / 0.140 & 0.352 / 0.172 & 0.325 / 0.162 & 0.342 / 0.200 & \strokeBold{0.475} / \strokeBold{0.275} \\
 &  & SD3 & 0.274 / 0.161 & 0.351 / 0.188 & 0.314 / 0.154 & 0.346 / 0.203 & \strokeBold{0.461} / \strokeBold{0.268} \\ \cmidrule(l){2-8} 
 &  & PP6 & 0.007 / 0.001 & 0.354 / 0.201 & 0.130 / 0.073 & 0.376 / 0.227 & \strokeBold{0.452} / \strokeBold{0.266} \\
 & PP8 & SD2 & 0.229 / 0.110 & 0.347 / 0.198 & 0.250 / 0.144 & 0.354 / 0.209 & \strokeBold{0.475} / \strokeBold{0.275} \\
\multirow{2}{*}{CoBEVT} &  & SD3 & 0.215 / 0.117 & 0.349 / 0.199 & 0.280 / 0.162 & 0.352 / 0.210 & \strokeBold{0.461} / \strokeBold{0.268} \\ \cmidrule(l){2-8} 
 &  & PP6 & 0.004 / 0.001 & 0.330 / 0.174 & 0.242 / 0.164 & 0.352 / 0.205 & \strokeBold{0.452} / \strokeBold{0.266} \\
 & SD2 & PP8 & 0.009 / 0.001 & 0.303 / 0.157 & 0.187 / 0.107 & 0.350 / 0.202 & \strokeBold{0.458} / \strokeBold{0.261} \\
 &  & SD3 & 0.267 / 0.166 & 0.374 / 0.211 & 0.398 / 0.228 & 0.278 / 0.121 & \strokeBold{0.461} / \strokeBold{0.268} \\ \cmidrule(l){2-8} 
 &  & PP6 & 0.007 / 0.001 & 0.263 / 0.078 & 0.269 / 0.150 & 0.346 / 0.201 & \strokeBold{0.452} / \strokeBold{0.266} \\
 & SD3 & PP8 & 0.013 / 0.001 & 0.317 / 0.174 & 0.151 / 0.072 & 0.377 / 0.224 & \strokeBold{0.458} / \strokeBold{0.261} \\
 &  & SD2 & 0.244 / 0.144 & 0.367 / 0.210 & 0.389 / 0.200 & 0.314 / 0.153 & \strokeBold{0.475} / \strokeBold{0.275} \\ \midrule
 &  & PP8 & 0.395 / 0.248 & 0.372 / 0.238 & 0.435 / 0.300 & 0.387 / 0.245 & \strokeBold{0.477} / \strokeBold{0.322} \\
 & PP6 & SD2 & 0.444 / 0.265 & 0.441 / 0.274 & 0.399 / 0.244 & 0.378 / 0.225 & \strokeBold{0.498} / \strokeBold{0.330} \\
 &  & SD3 & 0.428 / 0.289 & 0.450 / 0.284 & 0.409 / 0.250 & 0.379 / 0.216 & \strokeBold{0.485} / \strokeBold{0.322} \\ \cmidrule(l){2-8} 
 &  & PP6 & 0.287 / 0.172 & 0.446 / 0.259 & 0.442 / 0.306 & 0.375 / 0.225 & \strokeBold{0.482} / \strokeBold{0.328} \\
 & PP8 & SD2 & 0.425 / 0.265 & 0.450 / 0.268 & 0.414 / 0.277 & 0.368 / 0.218 & \strokeBold{0.498} / \strokeBold{0.330} \\
\multirow{2}{*}{V2X-ViT} &  & SD3 & 0.418 / 0.262 & 0.450 / 0.270 & 0.426 / 0.285 & 0.388 / 0.242 & \strokeBold{0.485} / \strokeBold{0.322} \\ \cmidrule(l){2-8} 
 &  & PP6 & 0.234 / 0.151 & 0.129 / 0.052 & 0.273 / 0.179 & 0.357 / 0.220 & \strokeBold{0.482} / \strokeBold{0.328} \\
 & SD2 & PP8 & 0.309 / 0.198 & 0.425 / 0.285 & 0.393 / 0.269 & 0.366 / 0.220 & \strokeBold{0.477} / \strokeBold{0.322} \\
 &  & SD3 & 0.407 / 0.283 & 0.435 / 0.297 & 0.344 / 0.218 & 0.380 / 0.225 & \strokeBold{0.485} / \strokeBold{0.322} \\ \cmidrule(l){2-8} 
 &  & PP6 & 0.376 / 0.223 & 0.285 / 0.178 & 0.034 / 0.019 & 0.371 / 0.221 & \strokeBold{0.482} / \strokeBold{0.328} \\
 & SD3 & PP8 & 0.344 / 0.205 & 0.419 / 0.257 & 0.042 / 0.023 & 0.372 / 0.224 & \strokeBold{0.477} / \strokeBold{0.322} \\
 &  & SD2 & 0.381 / 0.225 & 0.425 / 0.273 & 0.068 / 0.035 & 0.395 / 0.257 & \strokeBold{0.498} / \strokeBold{0.330} \\ \midrule
 &  & PP8 & 0.333 / 0.210 & 0.310 / 0.197 & 0.129 / 0.064 & 0.336 / 0.190 & \strokeBold{0.501} / \strokeBold{0.347} \\
 & PP6 & SD2 & 0.308 / 0.191 & 0.262 / 0.117 & 0.325 / 0.162 & 0.338 / 0.202 & \strokeBold{0.520} / \strokeBold{0.357} \\
 &  & SD3 & 0.315 / 0.201 & 0.333 / 0.221 & 0.314 / 0.154 & 0.339 / 0.207 & \strokeBold{0.516} / \strokeBold{0.355} \\ \cmidrule(l){2-8} 
 &  & PP6 & 0.343 / 0.218 & 0.147 / 0.039 & 0.130 / 0.073 & 0.372 / 0.229 & \strokeBold{0.503} / \strokeBold{0.352} \\
 & PP8 & SD2 & 0.317 / 0.206 & 0.403 / 0.282 & 0.250 / 0.144 & 0.370 / 0.235 & \strokeBold{0.520} / \strokeBold{0.357} \\
\multirow{2}{*}{ParCon} &  & SD3 & 0.341 / 0.220 & 0.401 / 0.282 & 0.280 / 0.162 & 0.359 / 0.222 & \strokeBold{0.516} / \strokeBold{0.355} \\ \cmidrule(l){2-8} 
 &  & PP6 & 0.387 / 0.250 & 0.358 / 0.227 & 0.242 / 0.164 & 0.358 / 0.220 & \strokeBold{0.503} / \strokeBold{0.352} \\
 & SD2 & PP8 & 0.368 / 0.228 & 0.418 / 0.289 & 0.187 / 0.107 & 0.333 / 0.204 & \strokeBold{0.501} / \strokeBold{0.347} \\
 &  & SD3 & 0.344 / 0.225 & 0.440 / 0.287 & 0.398 / 0.228 & 0.328 / 0.204 & \strokeBold{0.516} / \strokeBold{0.355} \\ \cmidrule(l){2-8} 
 &  & PP6 & 0.332 / 0.204 & 0.370 / 0.250 & 0.269 / 0.150 & 0.322 / 0.196 & \strokeBold{0.503} / \strokeBold{0.352} \\
 & SD3 & PP8 & 0.341 / 0.193 & 0.203 / 0.116 & 0.151 / 0.072 & 0.328 / 0.192 & \strokeBold{0.501} / \strokeBold{0.347} \\
 &  & SD2 & 0.309 / 0.174 & 0.414 / 0.274 & 0.389 / 0.200 & 0.315 / 0.186 & \strokeBold{0.520} / \strokeBold{0.357} \\ \bottomrule
\end{tabular}
\end{table}

\begin{table}[h!]
\caption{Zero-shot unseen evaluation on (\EGO+\AUXTWO) with LiDAR beam-count heterogeneity. Models trained/fine-tuned on (\EGO+\AUXONE) are directly evaluated on (\EGO+\AUXTWO) without additional training. To enable transfer from \AUXONE to \AUXTWO, the output representation of \AUXTWO\ is aligned to the auxiliary feature size used with \AUXONE. The best performance is highlighted in bold. In this setting, \EGO\ and \AUXONE\ use 128-beam LiDAR, whereas \AUXTWO\ uses 64-beam LiDAR.}
\label{tab:app_unseen-zero-shot_eval-LiDAR}
\centering
\footnotesize
\begin{tabular}{@{}ccc|ccccc@{}}
\toprule
Fusion & \multirow{2}{*}{\AUXONE} & \multirow{2}{*}{\AUXTWO} & E2E & MPDA & CodeFilling & GenComm & ALF (Ours) \\
Method &  &  & mAP@0.5/0.7 & mAP@0.5/0.7 & mAP@0.5/0.7 & mAP@0.5/0.7 & mAP@0.5/0.7 \\ \midrule
 & PP6 & PP6 & 0.071 /   0.023 & 0.378 / 0.261 & 0.201 / 0.143 & 0.431 / 0.279 & \strokeBold{0.446} / \strokeBold{0.299} \\
\multirow{2}{*}{AttnFuse} & PP8 & PP8 & 0.047 / 0.020 & 0.008 / 0.005 & 0.116 / 0.077 & 0.432 / 0.282 & \strokeBold{0.447} / \strokeBold{0.298} \\
 & SD2 & SD2 & 0.413 / 0.251 & 0.443 / 0.291 & 0.441 / 0.285 & 0.425 / 0.280 & \strokeBold{0.466} / \strokeBold{0.307} \\
 & SD3 & SD3 & 0.414 / 0.272 & 0.433 / 0.276 & 0.440 / 0.282 & 0.426 / 0.281 & \strokeBold{0.464} / \strokeBold{0.309} \\ \midrule
 & PP6 & PP6 & 0.007 / 0.001 & 0.175 / 0.041 & 0.232 / 0.106 & 0.339 / 0.201 & \strokeBold{0.426} / \strokeBold{0.238} \\
\multirow{2}{*}{CoBEVT} & PP8 & PP8 & 0.030 / 0.009 & 0.352 / 0.196 & 0.172 / 0.097 & 0.361 / 0.213 & \strokeBold{0.408} / \strokeBold{0.234} \\
 & SD2 & SD2 & 0.313 / 0.188 & 0.381 / 0.227 & 0.431 / 0.250 & 0.375 / 0.222 & \strokeBold{0.473} / \strokeBold{0.257} \\
 & SD3 & SD3 & 0.348 / 0.184 & 0.368 / 0.212 & 0.409 / 0.235 & 0.363 / 0.205 & \strokeBold{0.475} / \strokeBold{0.242} \\ \midrule
 & PP6 & PP6 & 0.369 / 0.214 & 0.364 / 0.221 & \strokeBold{0.448} / \strokeBold{0.307} & 0.386 / 0.246 & 0.412 / 0.291 \\
\multirow{2}{*}{V2X-ViT} & PP8 & PP8 & 0.391 / 0.252 & 0.360 / 0.212 & \strokeBold{0.443} / \strokeBold{0.308} & 0.385 / 0.223 & 0.410 / 0.271 \\
 & SD2 & SD2 & 0.431 / 0.287 & 0.437 / 0.293 & 0.424 / 0.278 & 0.389 / 0.238 & \strokeBold{0.441} / \strokeBold{0.324} \\
 & SD3 & SD3 & \strokeBold{0.437} / 0.278 & 0.430 / 0.277 & 0.398 / 0.246 & 0.352 / 0.206 & 0.433 / \strokeBold{0.318} \\ \midrule
 & PP6 & PP6 & 0.229 / 0.134 & 0.276 / 0.179 & 0.426 / 0.285 & 0.354 / 0.221 & \strokeBold{0.468} / \strokeBold{0.323} \\
\multirow{2}{*}{ParCon} & PP8 & PP8 & 0.362 / 0.224 & 0.416 / 0.273 & 0.429 / 0.282 & 0.351 / 0.213 & \strokeBold{0.464} / \strokeBold{0.319} \\
 & SD2 & SD2 & 0.409 / 0.258 & 0.442 / 0.284 & 0.440 / 0.290 & 0.332 / 0.199 & \strokeBold{0.499} / \strokeBold{0.349} \\
 & SD3 & SD3 & 0.390 / 0.248 & 0.433 / 0.291 & 0.436 / 0.276 & 0.346 / 0.208 & \strokeBold{0.488} / \strokeBold{0.336} \\ \bottomrule
\end{tabular}
\end{table}

\begin{figure}[h!]
    \centering
    \includegraphics[width=0.97\linewidth]{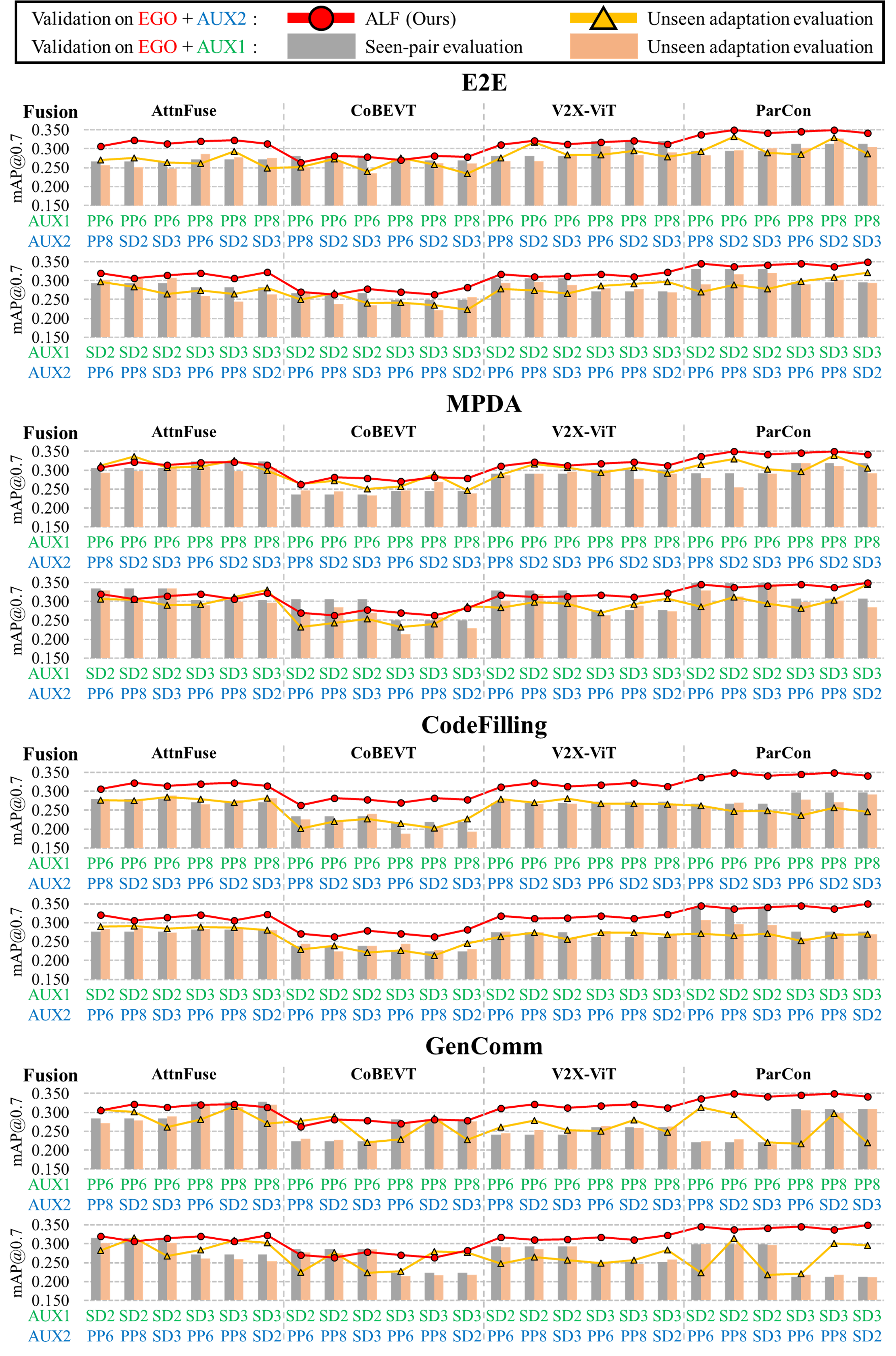}
    \vspace{-5pt}
    \caption{Comparison of mAP@0.7 under seen-pair and unseen-adaptation evaluation with encoder configuration heterogeneity. In this setting, \EGO, \AUXONE, and \AUXTWO\ are all equipped with 128-beam LiDARs.}
    \label{fig:sec4_unseen_adaptation_evaluation}
\end{figure}

\subsection{Full Results of Unseen Adaptation Evaluation}
\label{app:subsec:unseen_adaptation_evaluation}
\begin{figure}[h!]
    \centering
    \includegraphics[width=0.97\linewidth]{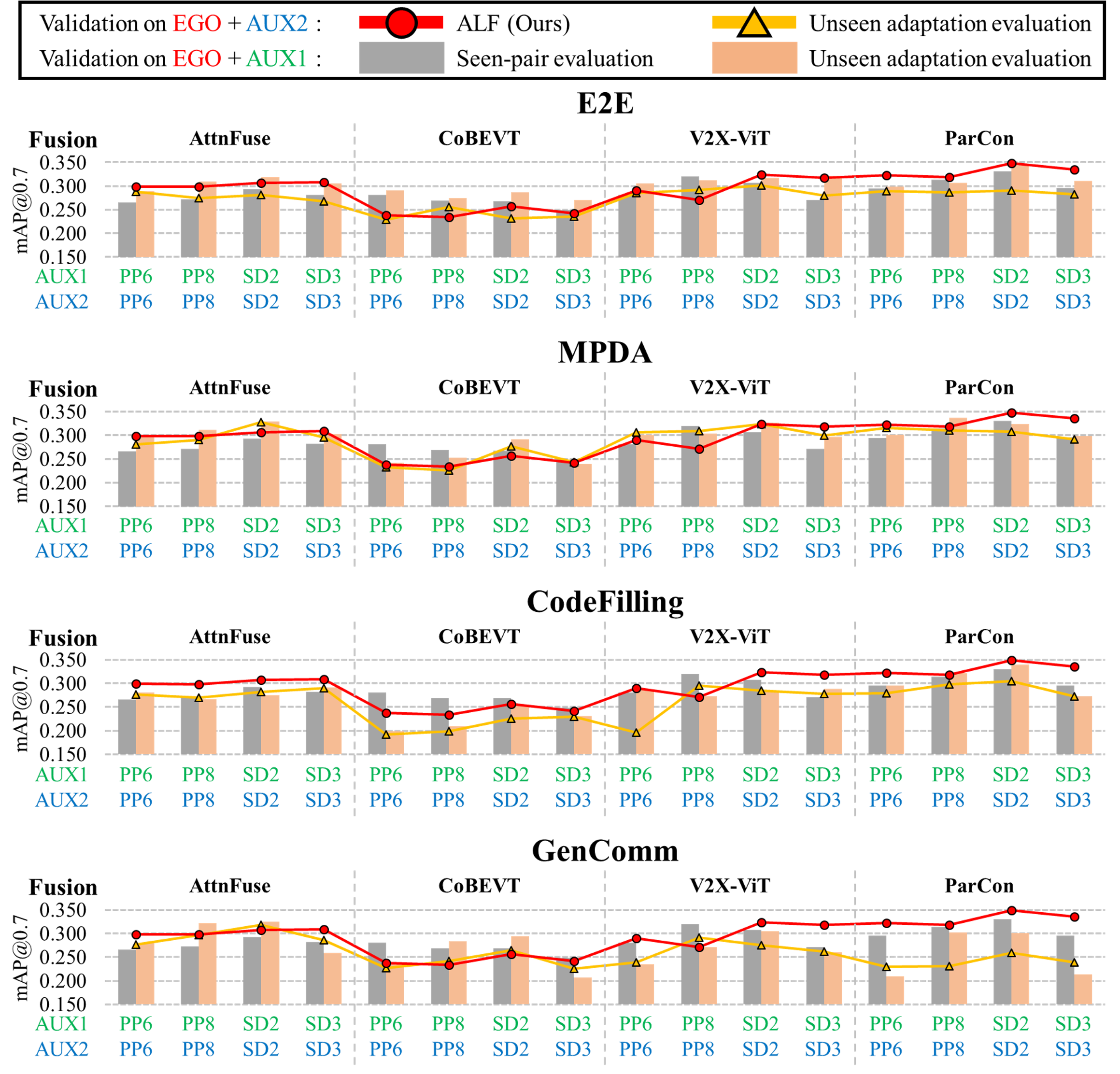}
    \vspace{-5pt}
    \caption{Comparison of mAP@0.7 under seen-pair and unseen-adaptation evaluation with LiDAR beam-count heterogeneity. In this setting, \EGO\ and \AUXONE\ use 128-beam LiDAR, whereas \AUXTWO\ uses 64-beam LiDAR.}
    \label{fig:sec4_unseen_adaptation_evaluation-lidar}
\end{figure}

Fig.~\ref{fig:sec4_unseen_adaptation_evaluation} and Fig.~\ref{fig:sec4_unseen_adaptation_evaluation-lidar} compare ALF with baseline methods in the unseen adaptation setting under encoder-configuration heterogeneity and LiDAR beam-count heterogeneity, respectively.

Under encoder-configuration heterogeneity, although ALF remains fixed after a single training stage, it still outperforms the adapted versions of E2E, MPDA, CodeFilling, and GenComm in 46, 43, 48, and 42 out of 48 \AUXONE--\AUXTWO--fusion-module combinations, respectively.

We further examine whether adapting to \AUXTWO\ preserves performance on the original seen configuration, \AUXONE. Across the 48 evaluated \AUXONE--\AUXTWO--fusion-module combinations, the unseen-adapted versions of E2E, MPDA, CodeFilling, and GenComm perform worse than their seen-pair counterparts in 33, 39, 29, and 32 cases, with maximum drops of 15.95\%, 14.70\%, 14.01\%, and 6.84\%, respectively.

Under LiDAR beam-count heterogeneity, ALF still outperforms the adapted versions of E2E, MPDA, CodeFilling, and GenComm in 14, 10, 15, and 12 out of 16 \AUXONE--\AUXTWO--fusion-module combinations, respectively. 

However, unlike the results under encoder-configuration heterogeneity, adaptation to the joint \EGO+\AUXONE+\AUXTWO\ setting is often beneficial in this case. Across the 16 evaluated \AUXONE--\AUXTWO--fusion-module combinations, the unseen-adapted versions of E2E, MPDA, CodeFilling, and GenComm perform worse than their seen-pair counterparts in only 2, 5, 9, and 11 cases, with maximum drops of 18.82\%, 14.55\%, 6.61\%, and 18.69\%, respectively. This trend suggests that, compared with the encoder-configuration heterogeneity setting, fine-tuning is more often beneficial under LiDAR beam-count heterogeneity, particularly for E2E and MPDA.

Tab.~\ref{tab:app_unseen-adaptation_eval_aux2} and Tab.~\ref{tab:app_unseen-adaptation_eval_aux1} report the full quantitative results for Fig.~\ref{fig:sec4_unseen_adaptation_evaluation}, while Tab.~\ref{tab:app_unseen-adaptation_eval_aux2-lidar} and Tab.~\ref{tab:app_unseen-adaptation_eval_aux1-lidar} report those for Fig.~\ref{fig:sec4_unseen_adaptation_evaluation-lidar}.

\begin{table}[t!]
\caption{Unseen adaptation evaluation on (\EGO+\AUXTWO) with encoder configuration heterogeneity. Models trained/fine-tuned on (\EGO+\AUXONE) are further trained/fine-tuned on (\EGO+\AUXONE+\AUXTWO) and then evaluated on (\EGO+\AUXTWO). The best performance is highlighted in bold. In this setting, \EGO, \AUXONE, and \AUXTWO\ are all equipped with 128-beam LiDARs.}
\label{tab:app_unseen-adaptation_eval_aux2}
\centering
\footnotesize
\begin{tabular}{@{}ccc|ccccc}
\toprule
Fusion & \multirow{2}{*}{AUX1} & \multirow{2}{*}{AUX2} & E2E & MPDA & CodeFilling & GenComm & ALF (Ours) \\
Method &  &  & mAP@0.5/0.7 & mAP@0.5/0.7 & mAP@0.5/0.7 & mAP@0.5/0.7 & mAP@0.5/0.7 \\ \midrule
 &  & PP8 & 0.433 / 0.271 & 0.479 / 0.312 & 0.432 / 0.277 & \strokeBold{0.489} / 0.307 & 0.472 / \strokeBold{0.317} \\
 & PP6 & SD2 & 0.460 / 0.276 & \strokeBold{0.511} / \strokeBold{0.337} & 0.426 / 0.275 & 0.482 / 0.301 & 0.496 / 0.327 \\
 &  & SD3 & 0.424 / 0.263 & 0.451 / 0.307 & 0.436 / 0.285 & 0.407 / 0.261 & \strokeBold{0.479} / \strokeBold{0.316} \\ \cmidrule(l){2-8} 
 &  & PP6 & 0.427 / 0.261 & 0.455 / 0.310 & 0.435 / 0.279 & 0.440 / 0.281 & \strokeBold{0.479} / \strokeBold{0.327} \\
 & PP8 & SD2 & 0.467 / 0.294 & \strokeBold{0.500} / 0.326 & 0.421 / 0.270 & 0.484 / 0.317 & 0.496 / \strokeBold{0.327} \\
\multirow{2}{*}{AttnFuse} &  & SD3 & 0.407 / 0.249 & 0.441 / 0.299 & 0.443 / 0.282 & 0.422 / 0.270 & \strokeBold{0.479} / \strokeBold{0.316} \\ \cmidrule(l){2-8} 
 &  & PP6 & 0.445 / 0.297 & 0.455 / 0.307 & 0.441 / 0.289 & 0.442 / 0.283 & \strokeBold{0.479} / \strokeBold{0.327} \\
 & SD2 & PP8 & 0.445 / 0.284 & 0.483 / 0.304 & 0.446 / 0.291 & \strokeBold{0.493} / 0.316 & 0.472 / \strokeBold{0.317} \\
 &  & SD3 & 0.433 / 0.265 & 0.444 / 0.290 & 0.440 / 0.284 & 0.419 / 0.268 & \strokeBold{0.479} / \strokeBold{0.316} \\ \cmidrule(l){2-8} 
 &  & PP6 & 0.427 / 0.274 & 0.458 / 0.291 & 0.442 / 0.289 & 0.446 / 0.283 & \strokeBold{0.479} / \strokeBold{0.327} \\
 & SD3 & PP8 & 0.440 / 0.264 & 0.482 / 0.312 & 0.445 / 0.288 & \strokeBold{0.488} / 0.310 & 0.472 / \strokeBold{0.317} \\
 &  & SD2 & 0.457 / 0.281 & \strokeBold{0.500} / \strokeBold{0.330} & 0.434 / 0.280 & 0.467 / 0.302 & 0.496 / 0.327 \\ \midrule
 &  & PP8 & 0.437 / 0.252 & \strokeBold{0.477} / 0.264 & 0.405 / 0.201 & 0.437 / \strokeBold{0.277} & 0.458 / 0.261 \\
 & PP6 & SD2 & \strokeBold{0.497} / 0.273 & 0.492 / 0.272 & 0.417 / 0.220 & 0.464 / \strokeBold{0.290} & 0.475 / 0.275 \\
 &  & SD3 & 0.431 / 0.240 & \strokeBold{0.464} / 0.250 & 0.400 / 0.226 & 0.392 / 0.221 & 0.461 / \strokeBold{0.268} \\ \cmidrule(l){2-8} 
 &  & PP6 & \strokeBold{0.468} / \strokeBold{0.276} & 0.437 / 0.258 & 0.405 / 0.215 & 0.402 / 0.230 & 0.452 / 0.266 \\
 & PP8 & SD2 & 0.477 / 0.259 & \strokeBold{0.486} / \strokeBold{0.289} & 0.405 / 0.203 & 0.464 / 0.286 & 0.475 / 0.275 \\
\multirow{2}{*}{CoBEVT} &  & SD3 & 0.432 / 0.234 & 0.427 / 0.247 & 0.420 / 0.227 & 0.399 / 0.228 & \strokeBold{0.461} / \strokeBold{0.268} \\ \cmidrule(l){2-8} 
 &  & PP6 & 0.415 / 0.249 & 0.432 / 0.232 & 0.408 / 0.229 & 0.407 / 0.224 & \strokeBold{0.452} / \strokeBold{0.266} \\
 & SD2 & PP8 & 0.450 / 0.267 & 0.440 / 0.244 & 0.421 / 0.239 & 0.444 / \strokeBold{0.277} & \strokeBold{0.458} / 0.261 \\
 &  & SD3 & 0.406 / 0.240 & 0.423 / 0.254 & 0.400 / 0.221 & 0.389 / 0.224 & \strokeBold{0.461} / \strokeBold{0.268} \\ \cmidrule(l){2-8} 
 &  & PP6 & 0.426 / 0.243 & 0.425 / 0.232 & 0.413 / 0.226 & 0.399 / 0.227 & \strokeBold{0.452} / \strokeBold{0.266} \\
 & SD3 & PP8 & 0.441 / 0.236 & 0.446 / 0.240 & 0.395 / 0.213 & 0.438 / \strokeBold{0.279} & \strokeBold{0.458} / 0.261 \\
 &  & SD2 & 0.428 / 0.224 & \strokeBold{0.500} / \strokeBold{0.288} & 0.421 / 0.245 & 0.467 / 0.277 & 0.475 / 0.275 \\ \midrule
 &  & PP8 & 0.472 / 0.276 & 0.460 / 0.288 & 0.438 / 0.277 & 0.432 / 0.261 & \strokeBold{0.477} / \strokeBold{0.322} \\
 & PP6 & SD2 & \strokeBold{0.502} / 0.317 & 0.492 / 0.316 & 0.416 / 0.269 & 0.455 / 0.279 & 0.498 / \strokeBold{0.330} \\
 &  & SD3 & 0.451 / 0.284 & 0.457 / 0.307 & 0.421 / 0.269 & 0.410 / 0.253 & \strokeBold{0.485} / \strokeBold{0.322} \\ \cmidrule(l){2-8} 
 &  & PP6 & 0.460 / 0.284 & 0.447 / 0.294 & 0.444 / 0.282 & 0.415 / 0.251 & \strokeBold{0.482} / \strokeBold{0.328} \\
 & PP8 & SD2 & 0.475 / 0.295 & \strokeBold{0.500} / 0.307 & 0.443 / 0.283 & 0.451 / 0.280 & 0.498 / \strokeBold{0.330} \\
\multirow{2}{*}{V2X-ViT} &  & SD3 & 0.428 / 0.279 & 0.442 / 0.292 & 0.412 / 0.263 & 0.415 / 0.248 & \strokeBold{0.485} / \strokeBold{0.322} \\ \cmidrule(l){2-8} 
 &  & PP6 & 0.449 / 0.278 & 0.452 / 0.283 & 0.430 / 0.271 & 0.412 / 0.247 & \strokeBold{0.482} / \strokeBold{0.328} \\
 & SD2 & PP8 & 0.439 / 0.275 & 0.461 / 0.298 & 0.432 / 0.274 & 0.430 / 0.264 & \strokeBold{0.477} / \strokeBold{0.322} \\
 &  & SD3 & 0.418 / 0.266 & 0.446 / 0.294 & 0.406 / 0.259 & 0.421 / 0.256 & \strokeBold{0.485} / \strokeBold{0.322} \\ \cmidrule(l){2-8} 
 &  & PP6 & 0.452 / 0.286 & 0.423 / 0.270 & 0.430 / 0.273 & 0.410 / 0.249 & \strokeBold{0.482} / \strokeBold{0.328} \\
 & SD3 & PP8 & 0.462 / 0.292 & 0.470 / 0.292 & 0.428 / 0.269 & 0.427 / 0.257 & \strokeBold{0.477} / \strokeBold{0.322} \\
 &  & SD2 & 0.471 / 0.297 & 0.488 / 0.307 & 0.412 / 0.261 & 0.461 / 0.284 & \strokeBold{0.498} / \strokeBold{0.330} \\ \midrule
 &  & PP8 & 0.450 / 0.294 & 0.479 / 0.315 & 0.440 / 0.261 & 0.489 / 0.314 & \strokeBold{0.501} / \strokeBold{0.347} \\
 & PP6 & SD2 & 0.518 / 0.332 & 0.515 / 0.329 & 0.407 / 0.246 & 0.495 / 0.294 & \strokeBold{0.520} / \strokeBold{0.357} \\
 &  & SD3 & 0.475 / 0.289 & 0.470 / 0.302 & 0.437 / 0.249 & 0.420 / 0.221 & \strokeBold{0.516} / \strokeBold{0.355} \\ \cmidrule(l){2-8} 
 &  & PP6 & 0.454 / 0.285 & 0.455 / 0.297 & 0.387 / 0.236 & 0.434 / 0.217 & \strokeBold{0.503} / \strokeBold{0.352} \\
 & PP8 & SD2 & \strokeBold{0.527} / 0.330 & 0.508 / 0.339 & 0.425 / 0.256 & 0.499 / 0.298 & 0.520 / \strokeBold{0.357} \\
\multirow{2}{*}{ParCon} &  & SD3 & 0.460 / 0.286 & 0.481 / 0.305 & 0.418 / 0.246 & 0.424 / 0.220 & \strokeBold{0.516} / \strokeBold{0.355} \\ \cmidrule(l){2-8} 
 &  & PP6 & 0.452 / 0.270 & 0.465 / 0.286 & 0.435 / 0.271 & 0.435 / 0.223 & \strokeBold{0.503} / \strokeBold{0.352} \\
 & SD2 & PP8 & 0.466 / 0.289 & 0.475 / 0.311 & 0.422 / 0.265 & 0.492 / 0.315 & \strokeBold{0.501} / \strokeBold{0.347} \\
 &  & SD3 & 0.455 / 0.278 & 0.454 / 0.294 & 0.425 / 0.271 & 0.417 / 0.218 & \strokeBold{0.516} / \strokeBold{0.355} \\ \cmidrule(l){2-8} 
 &  & PP6 & 0.447 / 0.298 & 0.431 / 0.282 & 0.411 / 0.252 & 0.431 / 0.221 & \strokeBold{0.503} / \strokeBold{0.352} \\
 & SD3 & PP8 & 0.476 / 0.309 & 0.471 / 0.303 & 0.430 / 0.267 & 0.478 / 0.301 & \strokeBold{0.501} / \strokeBold{0.347} \\
 &  & SD2 & 0.515 / 0.322 & 0.501 / 0.345 & 0.441 / 0.269 & 0.494 / 0.296 & \strokeBold{0.520} / \strokeBold{0.357} \\ \bottomrule
\end{tabular}
\end{table}

\begin{table}[t!]
\caption{Unseen adaptation evaluation on (\EGO+\AUXONE) with encoder configuration heterogeneity. Models trained/fine-tuned on (\EGO+\AUXONE) are further trained/fine-tuned on (\EGO+\AUXONE+\AUXTWO) and then evaluated on (\EGO+\AUXONE). The best performance is highlighted in bold. In this setting, \EGO, \AUXONE, and \AUXTWO\ are all equipped with 128-beam LiDARs.}
\label{tab:app_unseen-adaptation_eval_aux1}
\centering
\footnotesize
\begin{tabular}{@{}ccc|ccccc}
\toprule
Fusion & \multirow{2}{*}{AUX1} & \multirow{2}{*}{AUX2} & E2E & MPDA & CodeFilling & GenComm & ALF (Ours) \\
Method &  &  & mAP@0.5/0.7 & mAP@0.5/0.7 & mAP@0.5/0.7 & mAP@0.5/0.7 & mAP@0.5/0.7 \\ \midrule
 &  & PP8 & 0.410 / 0.257 & 0.456 / 0.293 & 0.430 / 0.276 & 0.438 / 0.273 & \strokeBold{0.472} / \strokeBold{0.317} \\
 & PP6 & SD2 & 0.400 / 0.252 & 0.452 / 0.298 & 0.433 / 0.277 & 0.448 / 0.279 & \strokeBold{0.496} / \strokeBold{0.327} \\
 &  & SD3 & 0.391 / 0.247 & 0.455 / 0.306 & 0.441 / 0.289 & 0.449 / 0.290 & \strokeBold{0.479} / \strokeBold{0.316} \\ \cmidrule(l){2-8} 
 &  & PP6 & 0.454 / 0.287 & 0.486 / 0.310 & 0.419 / 0.266 & \strokeBold{0.496} / 0.319 & 0.479 / \strokeBold{0.327} \\
 & PP8 & SD2 & 0.458 / 0.277 & 0.486 / 0.298 & 0.438 / 0.277 & 0.492 / 0.315 & \strokeBold{0.496} / \strokeBold{0.327} \\
\multirow{2}{*}{AttnFuse} &  & SD3 & 0.428 / 0.275 & 0.495 / 0.305 & 0.445 / 0.282 & \strokeBold{0.495} / \strokeBold{0.320} & 0.479 / 0.316 \\ \cmidrule(l){2-8} 
 &  & PP6 & 0.473 / 0.302 & \strokeBold{0.515} / \strokeBold{0.329} & 0.429 / 0.283 & 0.487 / 0.301 & 0.479 / 0.327 \\
 & SD2 & PP8 & 0.476 / 0.304 & \strokeBold{0.506} / 0.301 & 0.435 / 0.286 & 0.491 / 0.310 & 0.472 / \strokeBold{0.317} \\
 &  & SD3 & 0.476 / 0.308 & \strokeBold{0.511} / \strokeBold{0.334} & 0.424 / 0.273 & 0.488 / 0.303 & 0.479 / 0.316 \\ \cmidrule(l){2-8} 
 &  & PP6 & 0.445 / 0.260 & 0.454 / 0.291 & 0.434 / 0.283 & 0.418 / 0.260 & \strokeBold{0.479} / \strokeBold{0.327} \\
 & SD3 & PP8 & 0.451 / 0.245 & 0.448 / 0.299 & 0.439 / 0.285 & 0.409 / 0.259 & \strokeBold{0.472} / \strokeBold{0.317} \\
 &  & SD2 & 0.434 / 0.263 & 0.452 / 0.296 & 0.434 / 0.280 & 0.404 / 0.253 & \strokeBold{0.496} / \strokeBold{0.327} \\ \midrule
 &  & PP8 & 0.444 / 0.252 & \strokeBold{0.459} / 0.247 & 0.424 / 0.225 & 0.403 / 0.231 & 0.458 / \strokeBold{0.261} \\
 & PP6 & SD2 & 0.442 / 0.268 & 0.428 / 0.244 & 0.417 / 0.222 & 0.401 / 0.227 & \strokeBold{0.475} / \strokeBold{0.275} \\
 &  & SD3 & 0.424 / 0.236 & 0.447 / 0.233 & 0.423 / 0.241 & 0.403 / 0.225 & \strokeBold{0.461} / \strokeBold{0.268} \\ \cmidrule(l){2-8} 
 &  & PP6 & \strokeBold{0.472} / 0.270 & 0.452 / 0.247 & 0.400 / 0.188 & 0.433 / \strokeBold{0.271} & 0.452 / 0.266 \\
 & PP8 & SD2 & 0.437 / 0.262 & \strokeBold{0.475} / 0.269 & 0.419 / 0.204 & 0.445 / \strokeBold{0.278} & 0.475 / 0.275 \\
\multirow{2}{*}{CoBEVT} &  & SD3 & \strokeBold{0.474} / 0.261 & 0.465 / 0.238 & 0.397 / 0.193 & 0.442 / \strokeBold{0.276} & 0.461 / 0.268 \\ \cmidrule(l){2-8} 
 &  & PP6 & 0.475 / 0.267 & \strokeBold{0.502} / 0.269 & 0.443 / 0.245 & 0.470 / \strokeBold{0.277} & 0.452 / 0.266 \\
 & SD2 & PP8 & 0.465 / 0.238 & \strokeBold{0.492} / \strokeBold{0.284} & 0.405 / 0.224 & 0.476 / 0.275 & 0.458 / 0.261 \\
 &  & SD3 & 0.437 / 0.235 & \strokeBold{0.474} / 0.270 & 0.423 / 0.239 & 0.470 / \strokeBold{0.285} & 0.461 / 0.268 \\ \cmidrule(l){2-8} 
 &  & PP6 & 0.447 / 0.238 & 0.423 / 0.213 & 0.407 / 0.244 & 0.390 / 0.215 & \strokeBold{0.452} / \strokeBold{0.266} \\
 & SD3 & PP8 & 0.436 / 0.221 & 0.430 / 0.258 & 0.398 / 0.226 & 0.390 / 0.217 & \strokeBold{0.458} / \strokeBold{0.261} \\
 &  & SD2 & 0.450 / 0.256 & 0.445 / 0.230 & 0.414 / 0.230 & 0.395 / 0.217 & \strokeBold{0.475} / \strokeBold{0.275} \\ \midrule
 &  & PP8 & 0.464 / 0.268 & 0.433 / 0.287 & 0.415 / 0.273 & 0.406 / 0.245 & \strokeBold{0.477} / \strokeBold{0.322} \\
 & PP6 & SD2 & 0.425 / 0.268 & 0.443 / 0.291 & 0.411 / 0.267 & 0.411 / 0.253 & \strokeBold{0.498} / \strokeBold{0.330} \\
 &  & SD3 & 0.445 / 0.285 & 0.471 / 0.296 & 0.415 / 0.268 & 0.411 / 0.251 & \strokeBold{0.485} / \strokeBold{0.322} \\ \cmidrule(l){2-8} 
 &  & PP6 & \strokeBold{0.486} / 0.307 & 0.472 / 0.301 & 0.425 / 0.263 & 0.430 / 0.264 & 0.482 / \strokeBold{0.328} \\
 & PP8 & SD2 & 0.444 / 0.284 & 0.453 / 0.278 & 0.430 / 0.265 & 0.432 / 0.258 & \strokeBold{0.498} / \strokeBold{0.330} \\
\multirow{2}{*}{V2X-ViT} &  & SD3 & 0.449 / 0.290 & 0.462 / 0.291 & 0.438 / 0.266 & 0.440 / 0.263 & \strokeBold{0.485} / \strokeBold{0.322} \\ \cmidrule(l){2-8} 
 &  & PP6 & 0.483 / 0.295 & \strokeBold{0.491} / 0.303 & 0.434 / 0.276 & 0.469 / 0.290 & 0.482 / \strokeBold{0.328} \\
 & SD2 & PP8 & 0.480 / 0.297 & \strokeBold{0.499} / 0.320 & 0.435 / 0.275 & 0.465 / 0.287 & 0.477 / \strokeBold{0.322} \\
 &  & SD3 & 0.467 / 0.288 & \strokeBold{0.498} / 0.317 & 0.422 / 0.260 & 0.476 / 0.292 & 0.485 / \strokeBold{0.322} \\ \cmidrule(l){2-8} 
 &  & PP6 & 0.452 / 0.280 & 0.418 / 0.264 & 0.428 / 0.277 & 0.410 / 0.252 & \strokeBold{0.482} / \strokeBold{0.328} \\
 & SD3 & PP8 & 0.462 / 0.278 & 0.439 / 0.289 & 0.420 / 0.271 & 0.414 / 0.246 & \strokeBold{0.477} / \strokeBold{0.322} \\
 &  & SD2 & 0.432 / 0.269 & 0.431 / 0.274 & 0.414 / 0.271 & 0.419 / 0.258 & \strokeBold{0.498} / \strokeBold{0.330} \\ \midrule
 &  & PP8 & 0.444 / 0.282 & 0.436 / 0.278 & 0.442 / 0.257 & 0.437 / 0.223 & \strokeBold{0.501} / \strokeBold{0.347} \\
 & PP6 & SD2 & 0.465 / 0.296 & 0.423 / 0.255 & 0.431 / 0.269 & 0.438 / 0.229 & \strokeBold{0.520} / \strokeBold{0.357} \\
 &  & SD3 & 0.477 / 0.301 & 0.457 / 0.291 & 0.443 / 0.243 & 0.428 / 0.216 & \strokeBold{0.516} / \strokeBold{0.355} \\ \cmidrule(l){2-8} 
 &  & PP6 & 0.484 / 0.303 & 0.482 / 0.319 & 0.443 / 0.278 & 0.486 / 0.305 & \strokeBold{0.503} / \strokeBold{0.352} \\
 & PP8 & SD2 & 0.495 / 0.326 & 0.475 / 0.311 & 0.450 / 0.271 & 0.489 / 0.297 & \strokeBold{0.520} / \strokeBold{0.357} \\
\multirow{2}{*}{ParCon} &  & SD3 & 0.462 / 0.304 & 0.472 / 0.292 & 0.465 / 0.292 & 0.487 / 0.308 & \strokeBold{0.516} / \strokeBold{0.355} \\ \cmidrule(l){2-8} 
 &  & PP6 & 0.492 / 0.290 & \strokeBold{0.507} / 0.329 & 0.479 / 0.307 & 0.492 / 0.299 & 0.503 / \strokeBold{0.352} \\
 & SD2 & PP8 & 0.499 / 0.317 & 0.481 / 0.313 & 0.470 / 0.297 & \strokeBold{0.503} / 0.298 & 0.501 / \strokeBold{0.347} \\
 &  & SD3 & 0.501 / 0.320 & 0.509 / 0.345 & 0.463 / 0.293 & 0.496 / 0.296 & \strokeBold{0.516} / \strokeBold{0.355} \\ \cmidrule(l){2-8} 
 &  & PP6 & 0.458 / 0.290 & 0.455 / 0.298 & 0.418 / 0.258 & 0.411 / 0.218 & \strokeBold{0.503} / \strokeBold{0.352} \\
 & SD3 & PP8 & 0.471 / 0.301 & 0.458 / 0.303 & 0.432 / 0.273 & 0.418 / 0.218 & \strokeBold{0.501} / \strokeBold{0.347} \\
 &  & SD2 & 0.472 / 0.294 & 0.457 / 0.285 & 0.442 / 0.269 & 0.412 / 0.212 & \strokeBold{0.520} / \strokeBold{0.357} \\ \bottomrule
\end{tabular}
\end{table}

\begin{table}[t!]
\caption{Unseen adaptation evaluation on (\EGO+\AUXTWO) with LiDAR beam-count heterogeneity. Models trained/fine-tuned on (\EGO+\AUXONE) are further trained/fine-tuned on (\EGO+\AUXONE+\AUXTWO) and then evaluated on (\EGO+\AUXTWO). The best performance is highlighted in bold. In this setting, \EGO\ and \AUXONE\ use 128-beam LiDAR, whereas \AUXTWO\ uses 64-beam LiDAR.}
\label{tab:app_unseen-adaptation_eval_aux2-lidar}
\centering
\footnotesize
\begin{tabular}{@{}ccc|ccccc@{}}
\toprule
Fusion & \multirow{2}{*}{AUX1} & \multirow{2}{*}{AUX2} & E2E & MPDA & CodeFilling & GenComm & ALF (Ours) \\
Method &  &  & mAP@0.5/0.7 & mAP@0.5/0.7 & mAP@0.5/0.7 & mAP@0.5/0.7 & mAP@0.5/0.7 \\ \midrule
 & PP6 & PP6 & \strokeBold{0.459} /   0.288 & 0.450 / 0.281 & 0.429 / 0.277 & 0.439 / 0.276 & 0.446 / \strokeBold{0.299} \\
\multirow{2}{*}{AttnFuse} & PP8 & PP8 & 0.427 / 0.274 & 0.456 / 0.290 & 0.437 / 0.270 & \strokeBold{0.473} / 0.297 & 0.447 / \strokeBold{0.298} \\
 & SD2 & SD2 & 0.451 / 0.282 & \strokeBold{0.492} / \strokeBold{0.328} & 0.443 / 0.282 & 0.481 / 0.319 & 0.466 / 0.307 \\
 & SD3 & SD3 & 0.419 / 0.267 & \strokeBold{0.470} / 0.295 & 0.444 / 0.290 & 0.455 / 0.286 & 0.464 / \strokeBold{0.309} \\ \midrule
 & PP6 & PP6 & 0.408 / 0.229 & 0.408 / 0.233 & 0.375 / 0.192 & 0.381 / 0.227 & \strokeBold{0.412} / \strokeBold{0.238} \\
\multirow{2}{*}{CoBEVT} & PP8 & PP8 & \strokeBold{0.424} / \strokeBold{0.256} & 0.402 / 0.226 & 0.369 / 0.199 & 0.416 / 0.241 & 0.410 / 0.234 \\
 & SD2 & SD2 & 0.409 / 0.231 & \strokeBold{0.484} / \strokeBold{0.276} & 0.389 / 0.225 & 0.445 / 0.265 & 0.441 / 0.257 \\
 & SD3 & SD3 & 0.377 / 0.235 & 0.414 / \strokeBold{0.244} & 0.403 / 0.229 & 0.373 / 0.226 & \strokeBold{0.433} / 0.242 \\ \midrule
 & PP6 & PP6 & 0.436 / 0.285 & \strokeBold{0.468} / \strokeBold{0.307} & 0.338 / 0.196 & 0.386 / 0.239 & 0.426 / 0.291 \\
\multirow{2}{*}{V2X-ViT} & PP8 & PP8 & 0.460 / 0.292 & \strokeBold{0.462} / \strokeBold{0.309} & 0.453 / 0.295 & 0.442 / 0.291 & 0.408 / 0.271 \\
 & SD2 & SD2 & 0.479 / 0.302 & \strokeBold{0.490} / \strokeBold{0.324} & 0.435 / 0.285 & 0.437 / 0.275 & 0.473 / 0.324 \\
 & SD3 & SD3 & 0.436 / 0.281 & 0.456 / 0.300 & 0.416 / 0.277 & 0.405 / 0.262 & \strokeBold{0.475} / \strokeBold{0.318} \\ \midrule
 & PP6 & PP6 & 0.439 / 0.289 & \strokeBold{0.477} / 0.316 & 0.446 / 0.280 & 0.404 / 0.229 & 0.468 / \strokeBold{0.323} \\
\multirow{2}{*}{ParCon} & PP8 & PP8 & 0.444 / 0.287 & \strokeBold{0.475} / 0.310 & 0.456 / 0.298 & 0.444 / 0.231 & 0.464 / \strokeBold{0.319} \\
 & SD2 & SD2 & 0.467 / 0.291 & 0.485 / 0.307 & 0.475 / 0.305 & 0.451 / 0.259 & \strokeBold{0.499} / \strokeBold{0.349} \\
 & SD3 & SD3 & 0.430 / 0.283 & 0.446 / 0.291 & 0.432 / 0.273 & 0.406 / 0.239 & \strokeBold{0.488} / \strokeBold{0.336} \\ \bottomrule
\end{tabular}
\end{table}

\begin{table}[t!]
\caption{Unseen adaptation evaluation on (\EGO+\AUXONE) with LiDAR beam-count heterogeneity. Models trained/fine-tuned on (\EGO+\AUXONE) are further trained/fine-tuned on (\EGO+\AUXONE+\AUXTWO) and then evaluated on (\EGO+\AUXONE). The best performance is highlighted in bold. In this setting, \EGO, \AUXONE, and \AUXTWO\ are all equipped with 128-beam LiDARs.}
\label{tab:app_unseen-adaptation_eval_aux1-lidar}
\centering
\footnotesize
\begin{tabular}{@{}ccc|ccccc@{}}
\toprule
Fusion & \multirow{2}{*}{AUX1} & \multirow{2}{*}{AUX2} & E2E & MPDA & CodeFilling & GenComm & ALF (Ours) \\
Method &  &  & mAP@0.5/0.7 & mAP@0.5/0.7 & mAP@0.5/0.7 & mAP@0.5/0.7 & mAP@0.5/0.7 \\ \midrule
 & PP6 & PP6 & \strokeBold{0.455} /   0.290 & 0.453 / 0.298 & 0.439 / 0.280 & 0.449 / 0.279 & 0.446 / \strokeBold{0.299} \\
\multirow{2}{*}{AttnFuse} & PP8 & PP8 & 0.490 / 0.309 & 0.486 / 0.312 & 0.433 / 0.267 & \strokeBold{0.495} / \strokeBold{0.323} & 0.447 / 0.298 \\
 & SD2 & SD2 & 0.502 / 0.319 & \strokeBold{0.507} / \strokeBold{0.330} & 0.429 / 0.275 & 0.503 / 0.325 & 0.466 / 0.307 \\
 & SD3 & SD3 & 0.475 / 0.306 & \strokeBold{0.481} / 0.302 & 0.445 / 0.291 & 0.431 / 0.259 & 0.464 / \strokeBold{0.309} \\ \midrule
 & PP6 & PP6 & \strokeBold{0.461} / \strokeBold{0.291} & 0.412 / 0.242 & 0.389 / 0.199 & 0.402 / 0.239 & 0.412 / 0.238 \\
\multirow{2}{*}{CoBEVT} & PP8 & PP8 & \strokeBold{0.466} / 0.275 & 0.451 / 0.253 & 0.409 / 0.210 & 0.445 / \strokeBold{0.283} & 0.410 / 0.234 \\
 & SD2 & SD2 & 0.487 / 0.286 & \strokeBold{0.498} / 0.292 & 0.423 / 0.256 & 0.475 / \strokeBold{0.294} & 0.441 / 0.257 \\
 & SD3 & SD3 & \strokeBold{0.454} / \strokeBold{0.270} & 0.431 / 0.239 & 0.400 / 0.231 & 0.400 / 0.207 & 0.433 / 0.242 \\ \midrule
 & PP6 & PP6 & \strokeBold{0.463} / \strokeBold{0.305} & 0.447 / 0.301 & 0.444 / 0.287 & 0.406 / 0.235 & 0.426 / 0.291 \\
\multirow{2}{*}{V2X-ViT} & PP8 & PP8 & \strokeBold{0.479} / \strokeBold{0.312} & 0.470 / 0.304 & 0.436 / 0.273 & 0.440 / 0.271 & 0.408 / 0.271 \\
 & SD2 & SD2 & 0.502 / 0.318 & \strokeBold{0.507} / \strokeBold{0.326} & 0.448 / 0.282 & 0.476 / 0.304 & 0.473 / 0.324 \\
 & SD3 & SD3 & \strokeBold{0.479} / \strokeBold{0.322} & 0.454 / 0.297 & 0.435 / 0.289 & 0.412 / 0.260 & 0.475 / 0.318 \\ \midrule
 & PP6 & PP6 & 0.454 / 0.299 & \strokeBold{0.477} / 0.301 & 0.476 / 0.296 & 0.432 / 0.210 & 0.468 / \strokeBold{0.323} \\
\multirow{2}{*}{ParCon} & PP8 & PP8 & 0.469 / 0.307 & \strokeBold{0.517} / \strokeBold{0.337} & 0.497 / 0.322 & 0.489 / 0.303 & 0.464 / 0.319 \\
 & SD2 & SD2 & \strokeBold{0.543} / 0.346 & 0.508 / 0.324 & 0.512 / 0.340 & 0.506 / 0.301 & 0.499 / \strokeBold{0.349} \\
 & SD3 & SD3 & \strokeBold{0.492} / 0.311 & 0.465 / 0.298 & 0.433 / 0.273 & 0.414 / 0.213 & 0.488 / \strokeBold{0.336} \\ \bottomrule
\end{tabular}
\end{table}


\clearpage

\end{document}